\documentclass{article}

\usepackage{microtype}
\usepackage{graphicx}
\usepackage{subcaption}
\usepackage{booktabs} 
\usepackage{amsmath}
\usepackage{placeins}
\usepackage{amsthm}
\usepackage{array} 
\usepackage{changepage}
\usepackage{enumitem}

\usepackage{lscape}

\usepackage{hyperref}

\usepackage[accepted]{icml2019}
\usepackage{xcolor}

\icmltitlerunning{Robustness to Extraneous Variables Via Adaptive Feature Normalization}

\begin{document}

\twocolumn[
\icmltitle{Be Like Water: Robustness to Extraneous Variables\\ Via Adaptive Feature Normalization}



\icmlsetsymbol{equal}{*}

\begin{icmlauthorlist}
\icmlauthor{Aakash Kaku}{equal,cds}
\icmlauthor{Sreyas Mohan}{equal,cds}
\icmlauthor{Avinash Parnandi}{nyumc}
\icmlauthor{Heidi Schambra}{nyumc}
\icmlauthor{Carlos Fernandez-Granda}{cds,courant}

\end{icmlauthorlist}

\icmlaffiliation{cds}{Center for Data Science, New York University, NY, USA}
\icmlaffiliation{nyumc}{Department of Neurology, New York University School of Medicine, NY, USA}
\icmlaffiliation{courant}{Courant Institute of Mathematical Sciences, New York University, NY, USA}

\icmlcorrespondingauthor{Sreyas Mohan}{sm7582@nyu.edu}
\icmlcorrespondingauthor{Aakash Kaku}{ark576@nyu.edu}

\icmlkeywords{Machine Learning, ICML}

\vskip 0.3in
]



\printAffiliationsAndNotice{\icmlEqualContribution} 

\begin{abstract}
Extraneous variables are variables that are irrelevant for a certain task, but heavily affect the distribution of the available data. In this work, we show that the presence of such variables can degrade the performance of deep-learning models. We study three datasets where there is a strong influence of known extraneous variables: classification of upper-body movements in stroke patients, annotation of surgical activities, and recognition of corrupted images. Models trained with batch normalization learn features that are highly dependent on the extraneous variables. In batch normalization, the statistics used to normalize the features are learned from the training set and fixed at test time, which produces a mismatch in the presence of varying extraneous variables. We demonstrate that estimating the feature statistics adaptively during inference, as in instance normalization, addresses this issue, producing normalized features that are more robust to changes in the extraneous variables. This results in a significant gain in performance for different network architectures and choices of feature statistics.
\end{abstract}



\section{Introduction}

\def\nsp{\hspace*{0in}}
\begin{figure*}[ht]
\begin{adjustwidth}{-0.2in}{}
\def\f1ht{1.42in}
\centering 
\begin{tabular}{
>{\centering\arraybackslash}m{0.23\linewidth}>{\centering\arraybackslash}m{0.23\linewidth}>{\centering\arraybackslash}m{0.23\linewidth}>{\centering\arraybackslash}m{0.23\linewidth}}
\multicolumn{2}{c}{\footnotesize{CIFAR-10}} & \multicolumn{2}{c}{\footnotesize{Fashion-MNIST}}\\
\includegraphics[height=\f1ht]{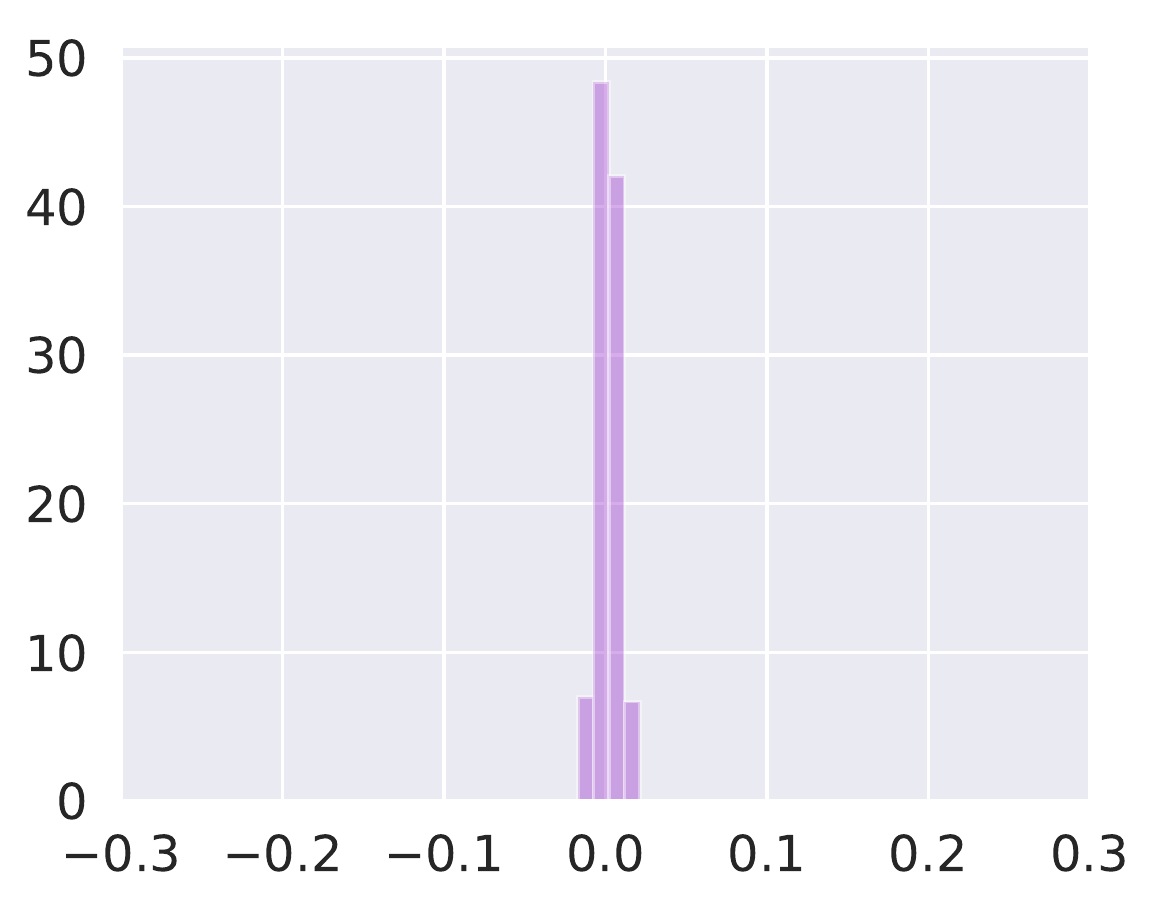} &
 \nsp \includegraphics[height=\f1ht]{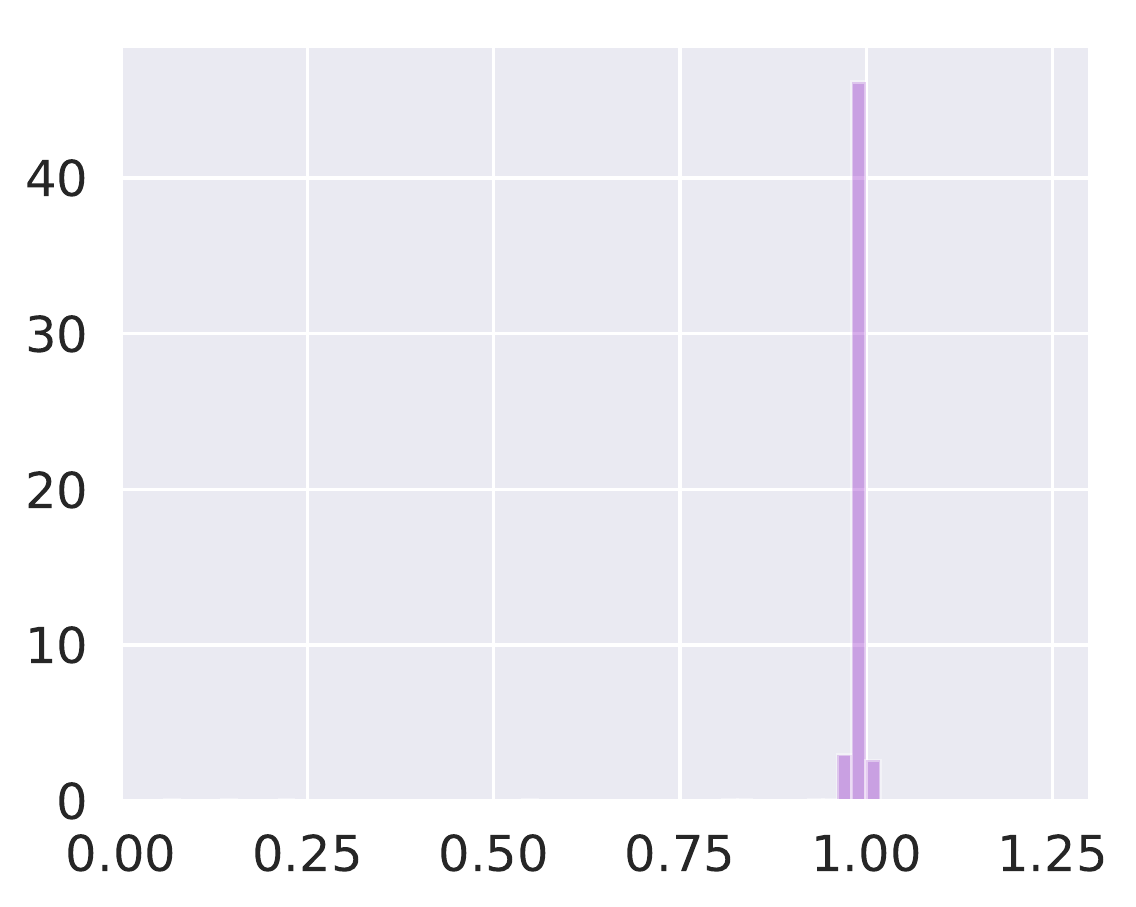}\nsp &
  \nsp\includegraphics[height=\f1ht]{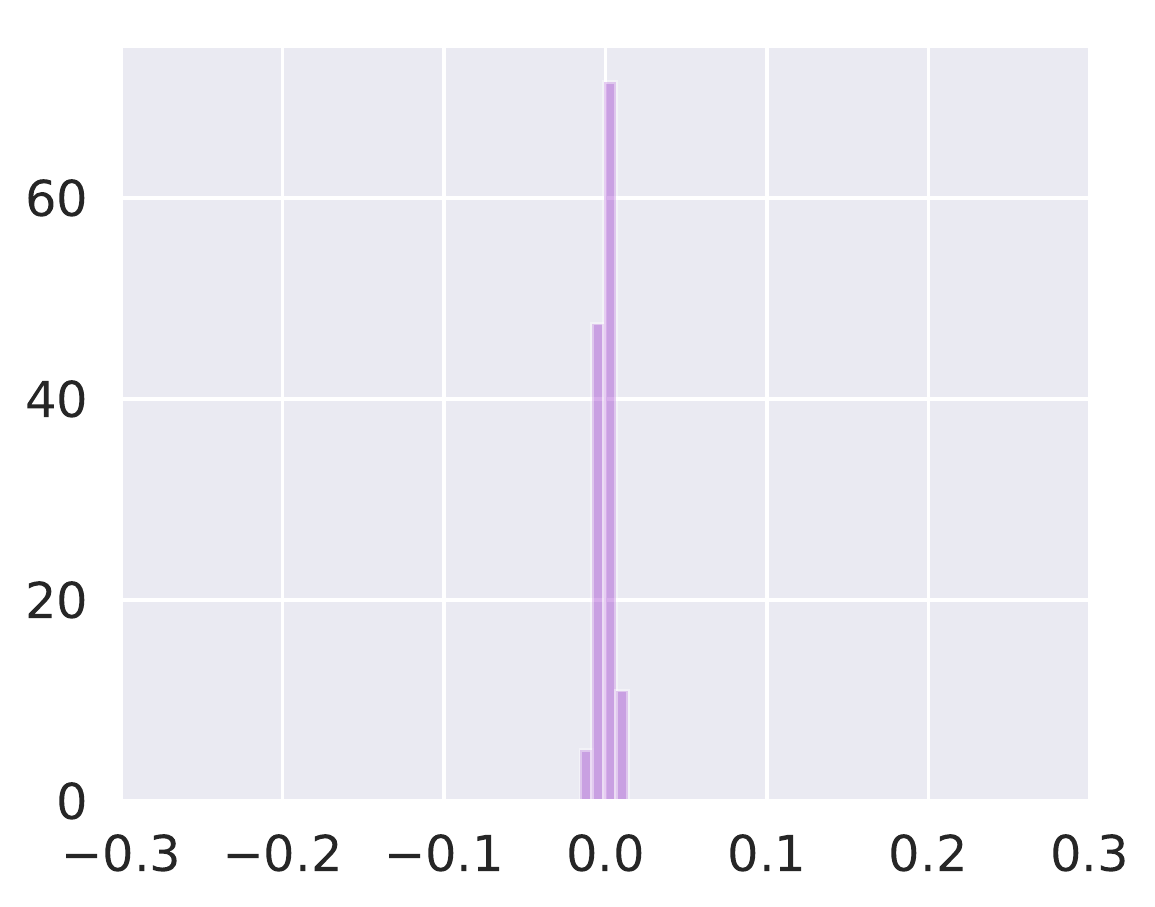} &
  \nsp\includegraphics[height=\f1ht]{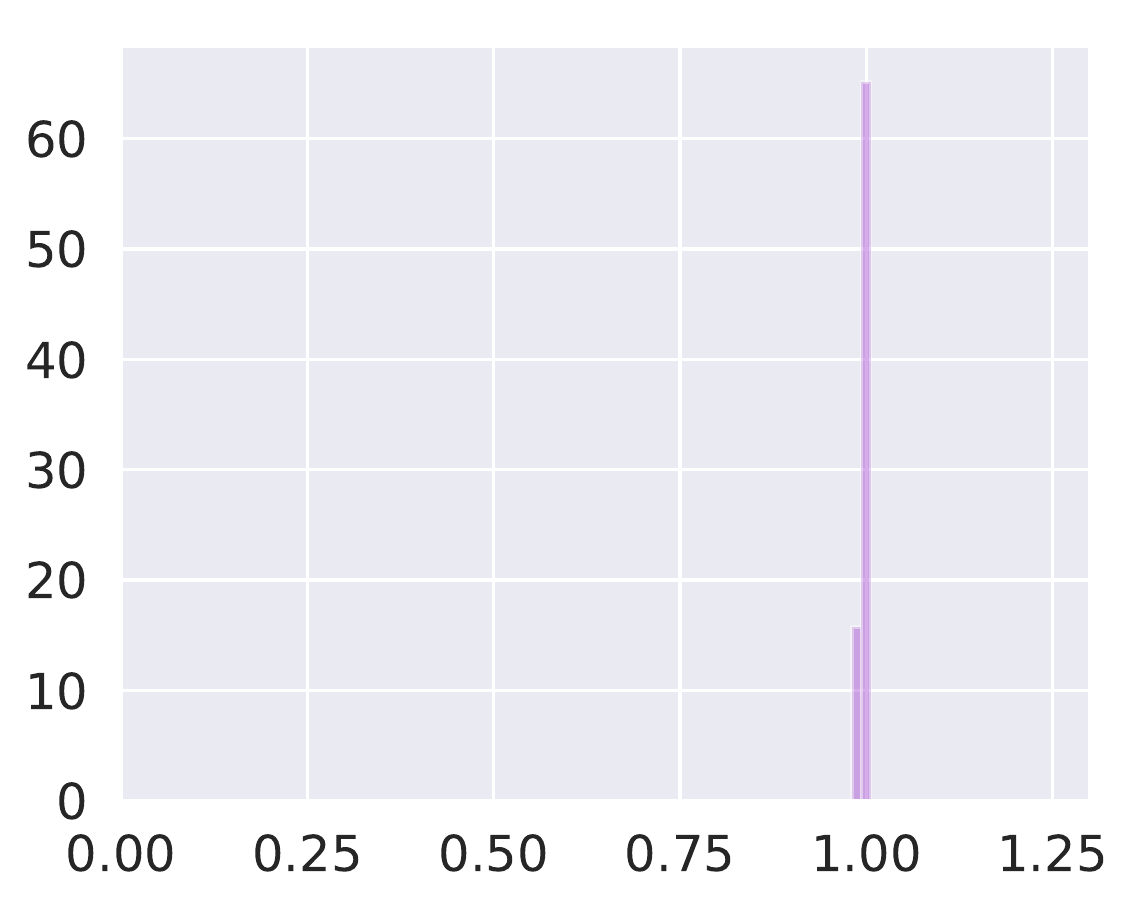} \\
  \footnotesize{Mean} \vspace{0.1cm} & \footnotesize{Standard Deviation} \vspace{0.1cm} & \footnotesize{Mean} \vspace{0.1cm} & \footnotesize{Standard Deviation} \vspace{0.1cm} \\
  \end{tabular}\\[-0.5ex]
\end{adjustwidth}
\caption{Histograms of the mean and standard deviations of normalized intermediate features each corresponding to a separate convolutional filter for the CIFAR-10 and Fashion-MNIST datasets. The normalization is carried out on test data using statistics estimated from the training data. The histogram of means concentrates around zero, whereas the histogram of standard deviations concentrates around one, showing that the non-adaptive normalization scheme is successful in normalizing the held-out data.}
\label{fig:bn_cifar10_fmnist}
\end{figure*}

\def\nsp{\hspace*{0in}}
\begin{figure*}[t]
\def\f1ht{1.65in}
\centering 
\begin{tabular}{
>{\centering\arraybackslash}m{0.06\linewidth}>{\centering\arraybackslash}m{0.28\linewidth}>{\centering\arraybackslash}m{0.28\linewidth}>{\centering\arraybackslash}m{0.28\linewidth}}
 &\footnotesize{CIFAR-10-C} \vspace{0.1cm} & \footnotesize{JIGSAWS} \vspace{0.1cm} & \footnotesize{StrokeRehab} \vspace{0.1cm} \\
  \scriptsize{Non Adaptive} & \nsp\includegraphics[height=\f1ht]{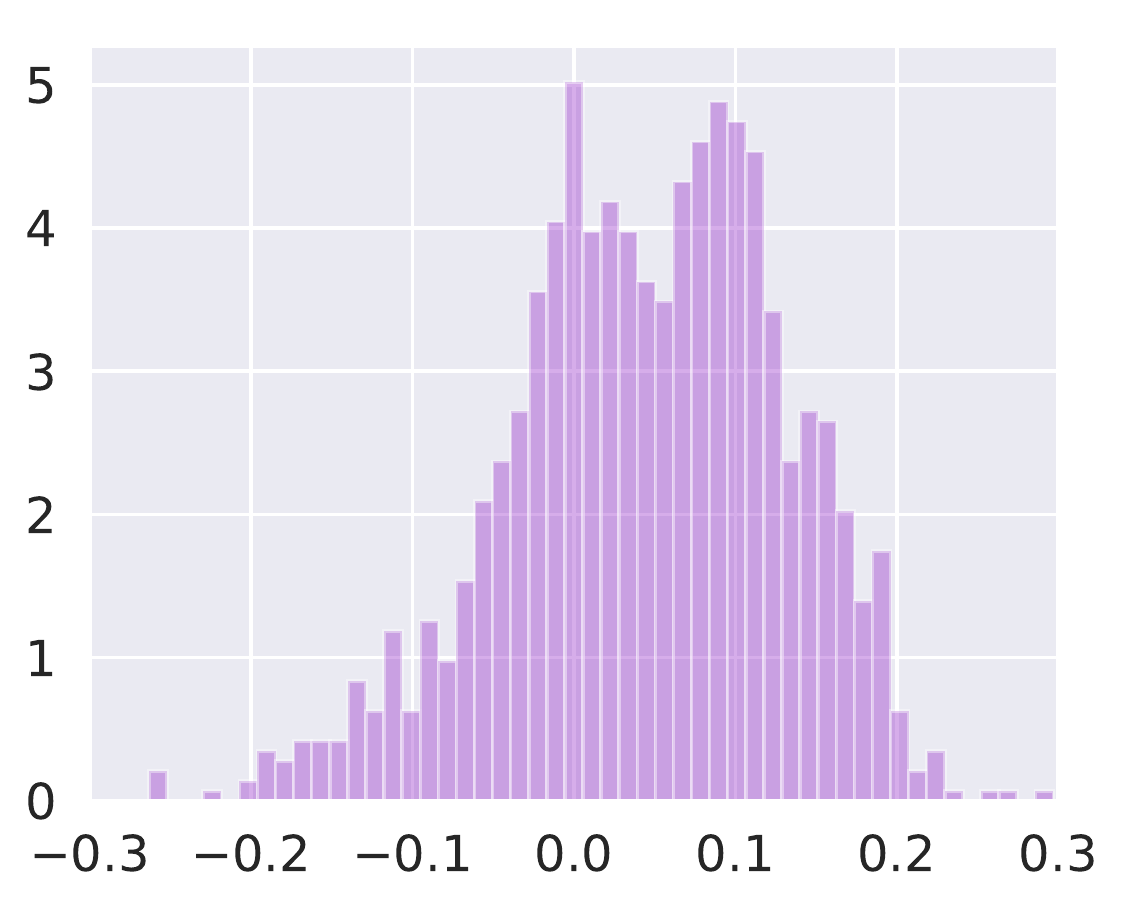}\nsp &
  \nsp\includegraphics[height=\f1ht]{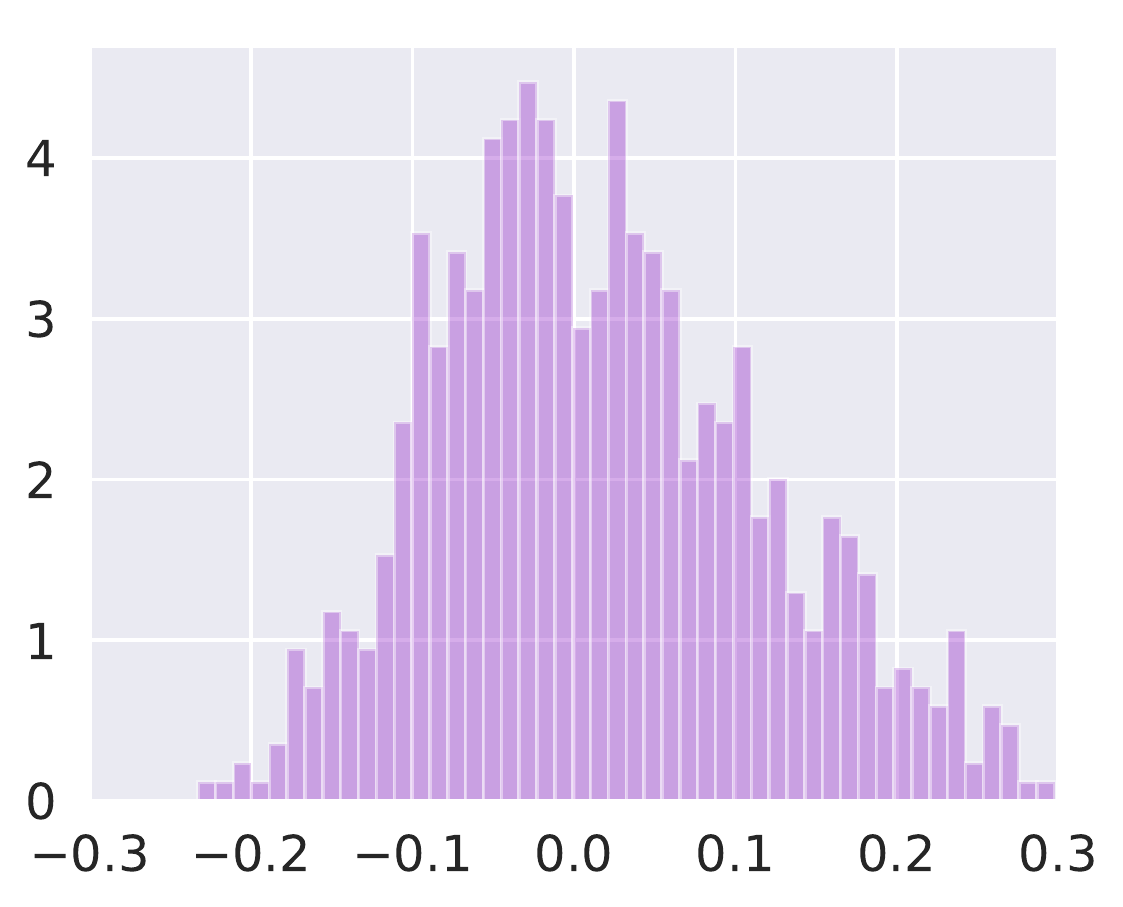}\nsp &
  \nsp\includegraphics[height=\f1ht]{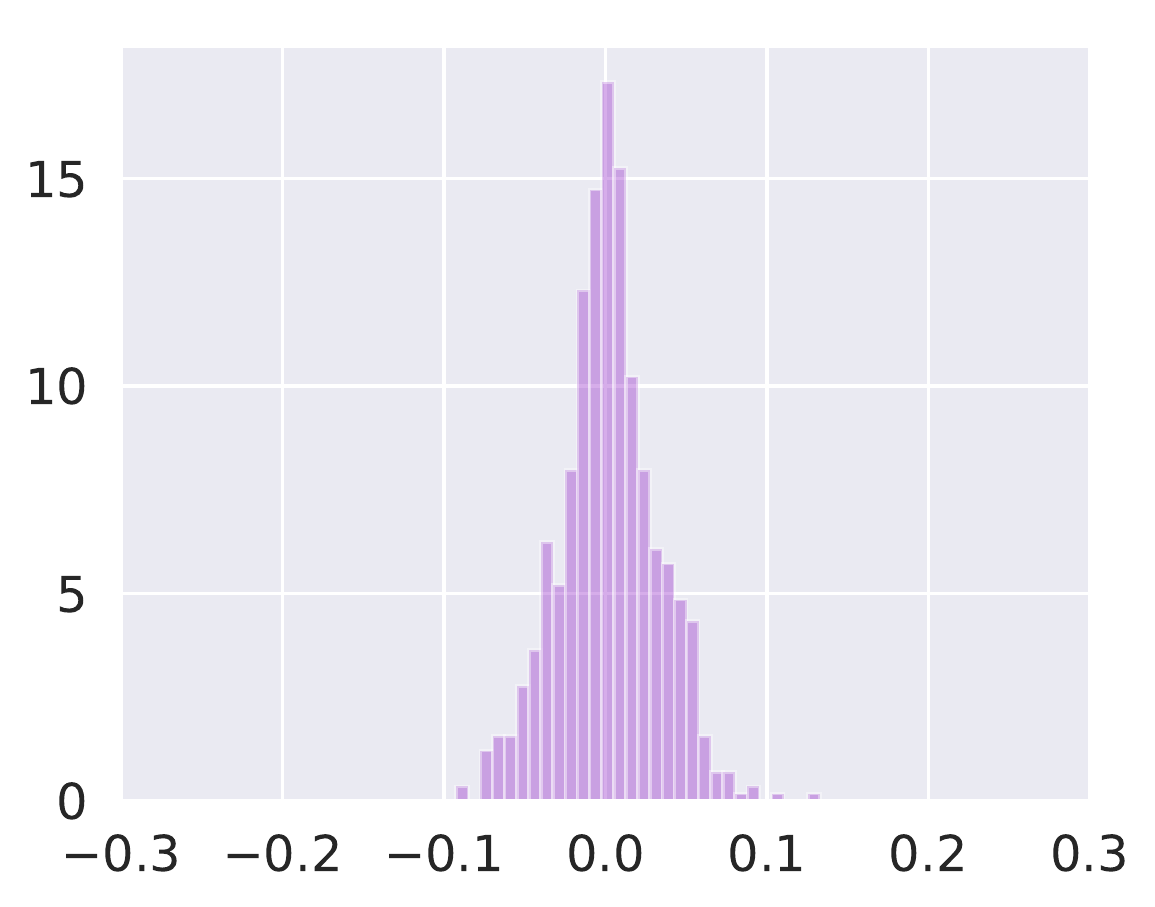} \\
  \scriptsize{Adaptive} & \nsp\includegraphics[height=\f1ht]{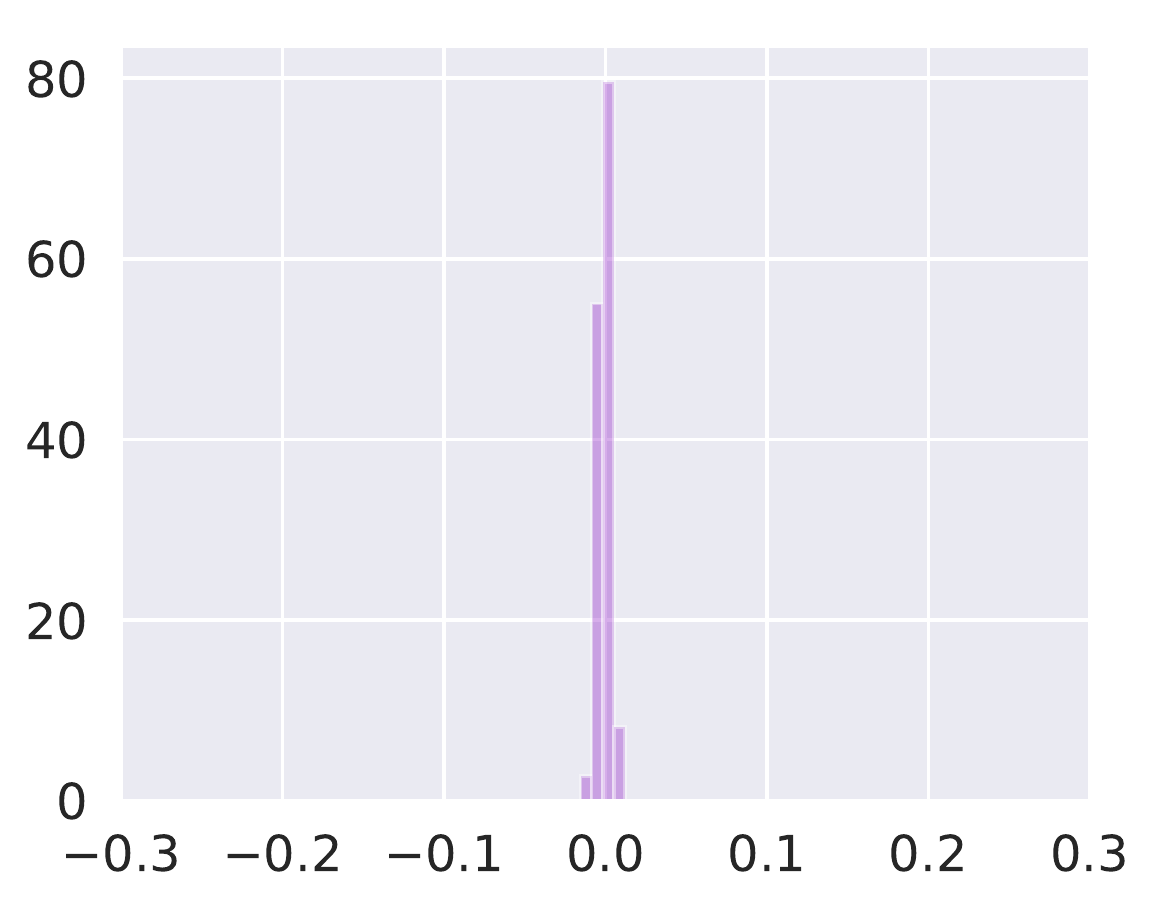}\nsp &
  \nsp\includegraphics[height=\f1ht]{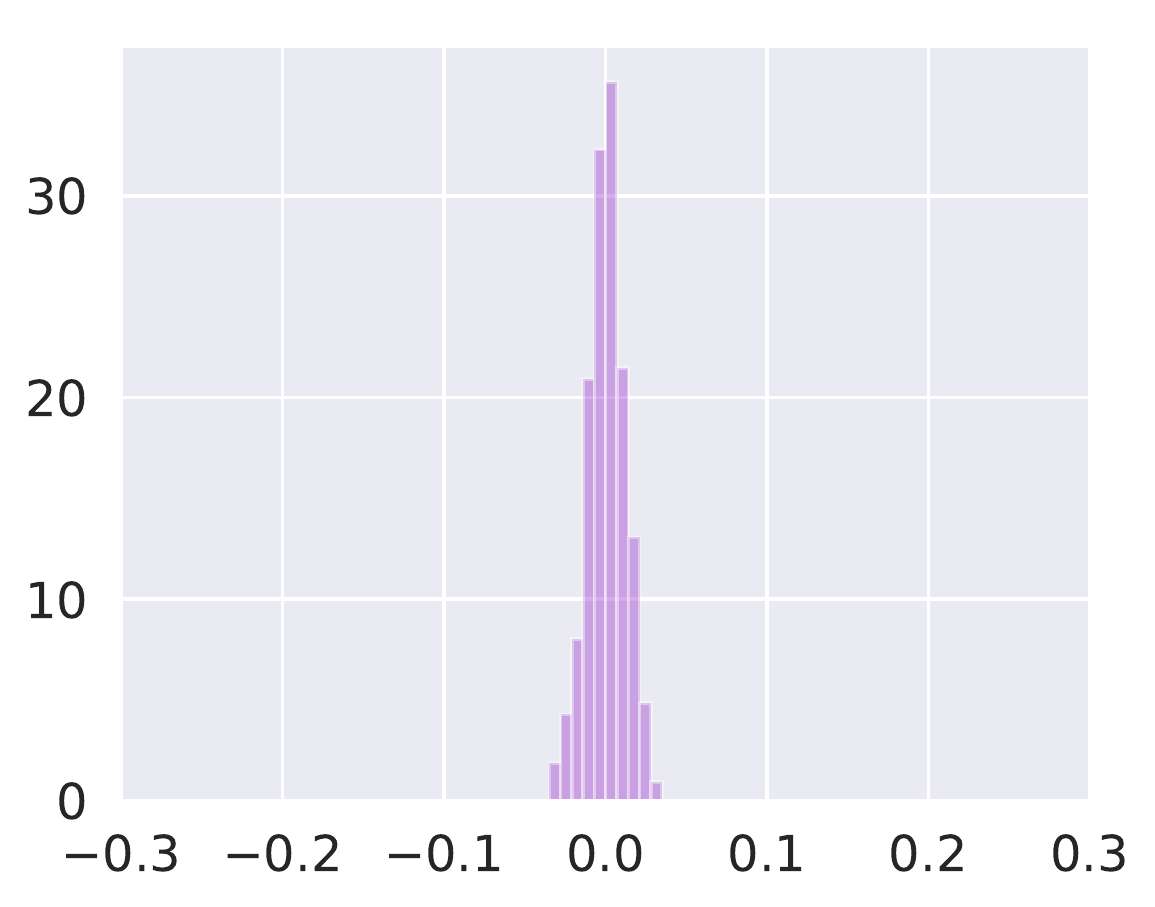}\nsp &
  \nsp\includegraphics[height=\f1ht]{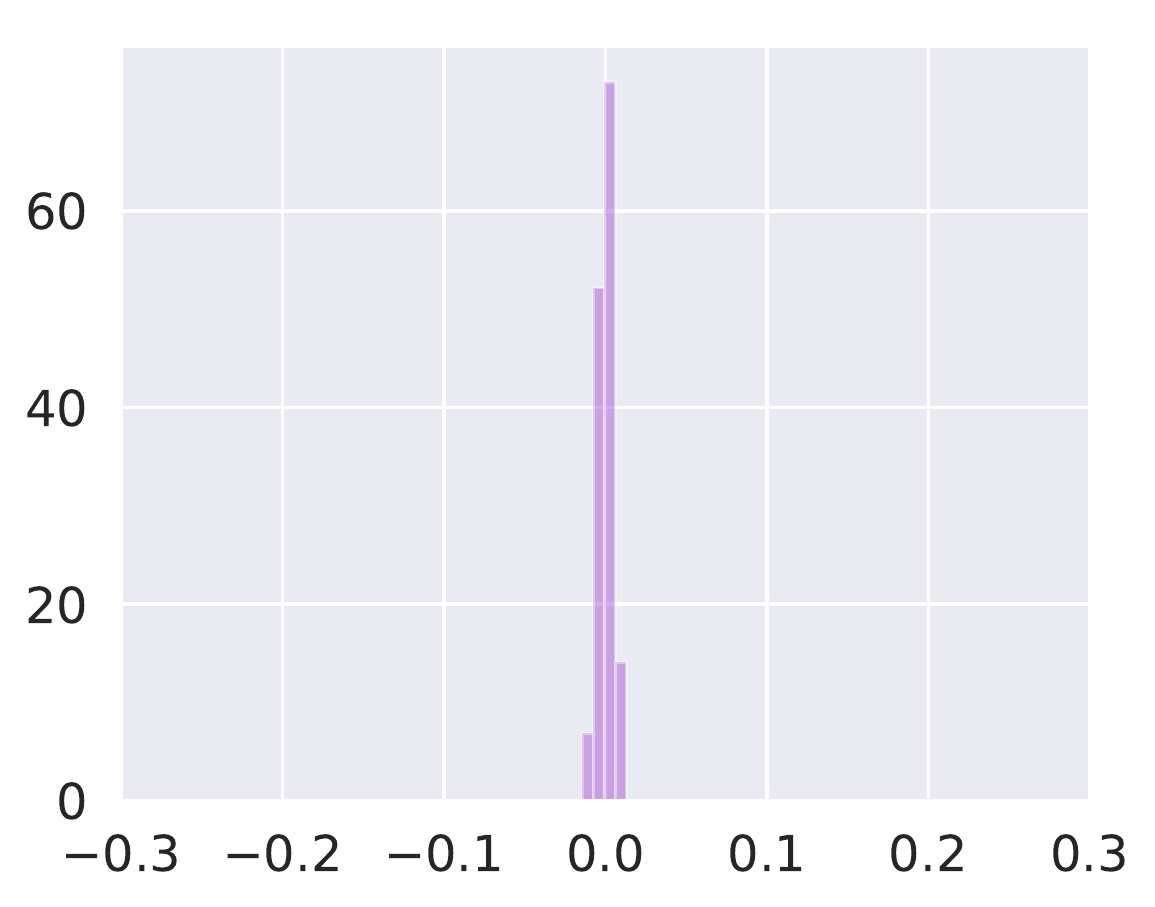} \\
  \end{tabular}\\[-0.5ex]
\caption{Histograms of the mean of normalized intermediate features each corresponding to a separate convolutional filter for the CIFAR-10-C, JIGSAWS, and StrokeRehab datasets. On the top row, the normalization is carried out on test data using statistics estimated from the training data (i.e. non-adaptively). The histogram of means does not concentrate around zero, as opposed to the mean histogram in Figure~\ref{fig:bn_cifar10_fmnist}. This indicates that the non-adaptive normalization scheme is not successful in normalizing the held-out data. On the bottom row, the normalization is carried out adaptively, using batches where the value of the extraneous variable is fixed. This results in a histogram that does concentrate around zero, demonstrating that adaptive normalization succeeds in normalizing the features correctly.}
\label{fig:hist_mean_ext}
\end{figure*}

\def\nsp{\hspace*{0in}}
\begin{figure*}[t]
\def\f1ht{1.65in}
\centering 
\begin{tabular}{
>{\centering\arraybackslash}m{0.06\linewidth}>{\centering\arraybackslash}m{0.28\linewidth}>{\centering\arraybackslash}m{0.28\linewidth}>{\centering\arraybackslash}m{0.28\linewidth}}
 &\footnotesize{CIFAR-10C} \vspace{0.1cm} & \footnotesize{JIGSAWS} \vspace{0.1cm} & \footnotesize{StrokeRehab} \vspace{0.1cm} \\
  \scriptsize{Non Adaptive} & \nsp\includegraphics[height=\f1ht]{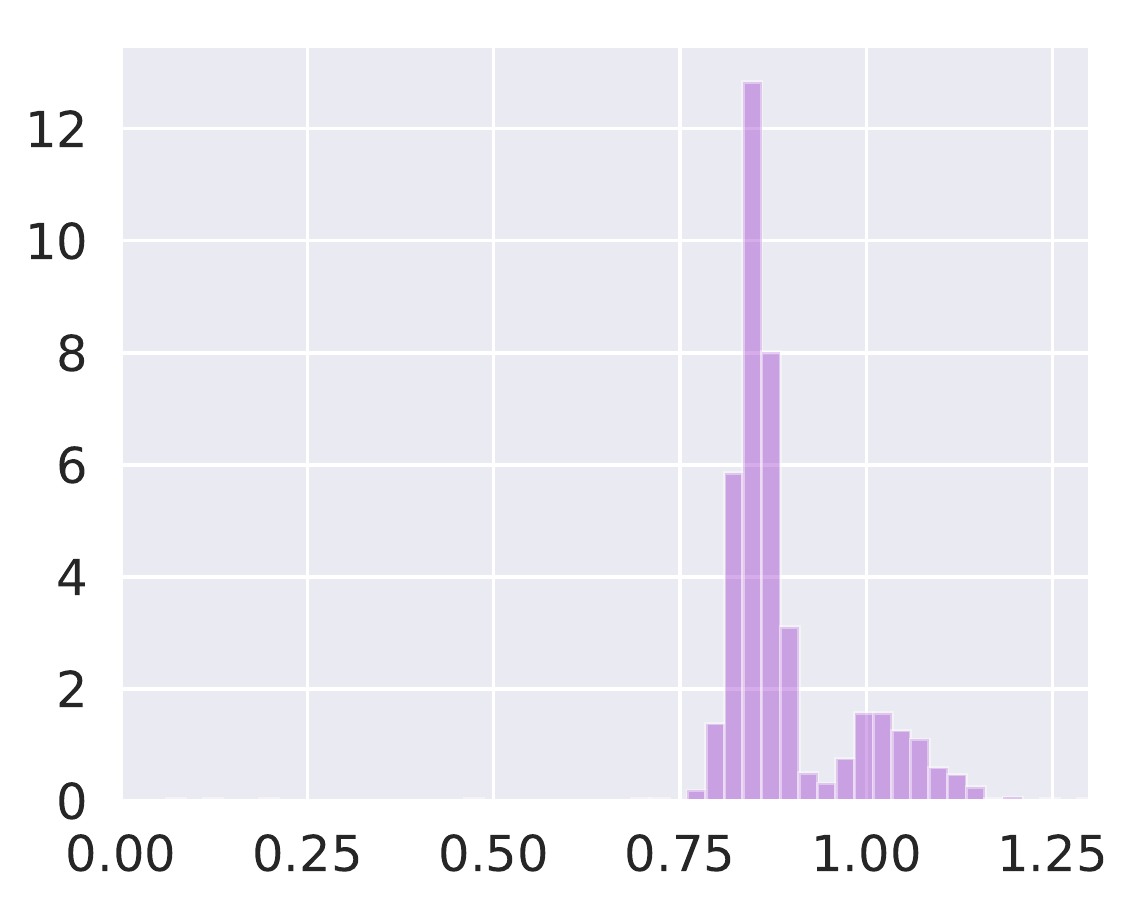}\nsp &
  \nsp\includegraphics[height=\f1ht]{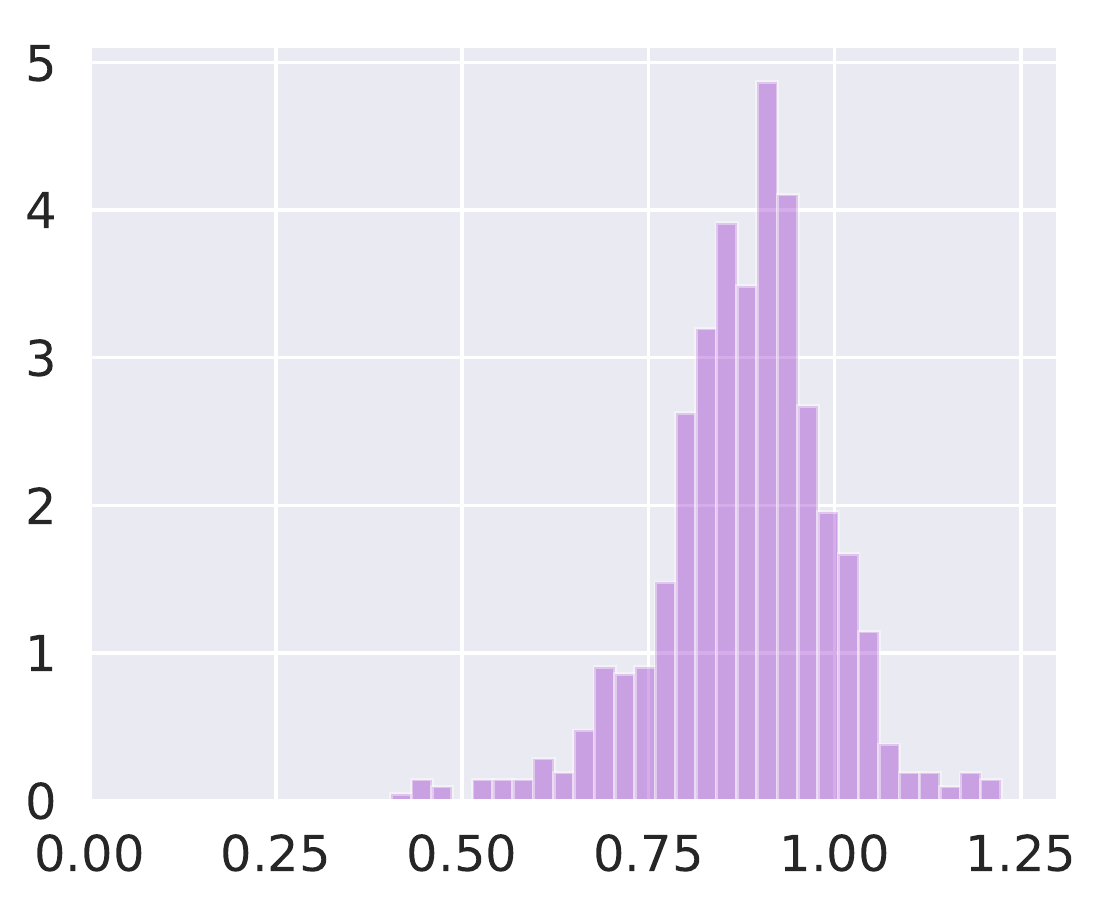}\nsp &
  \nsp\includegraphics[height=\f1ht]{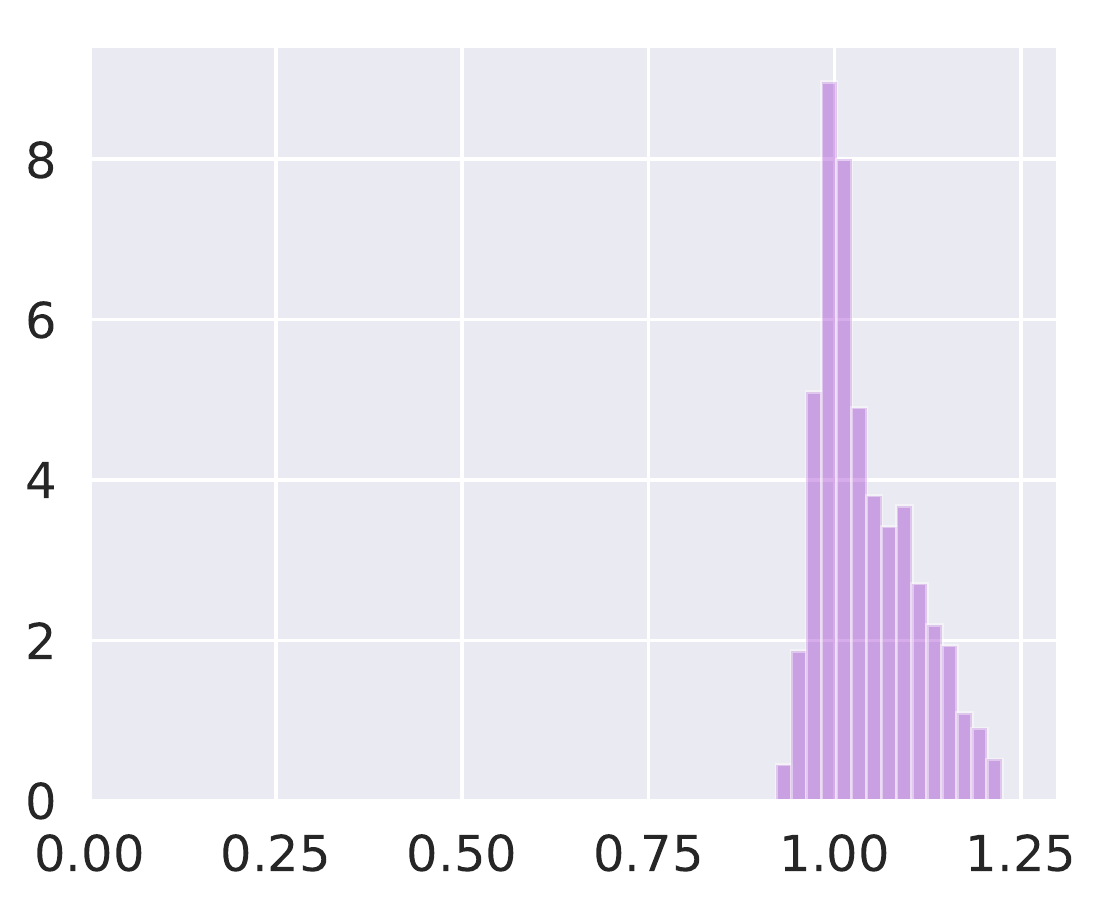} \\
  \scriptsize{Adaptive} & \nsp\includegraphics[height=\f1ht]{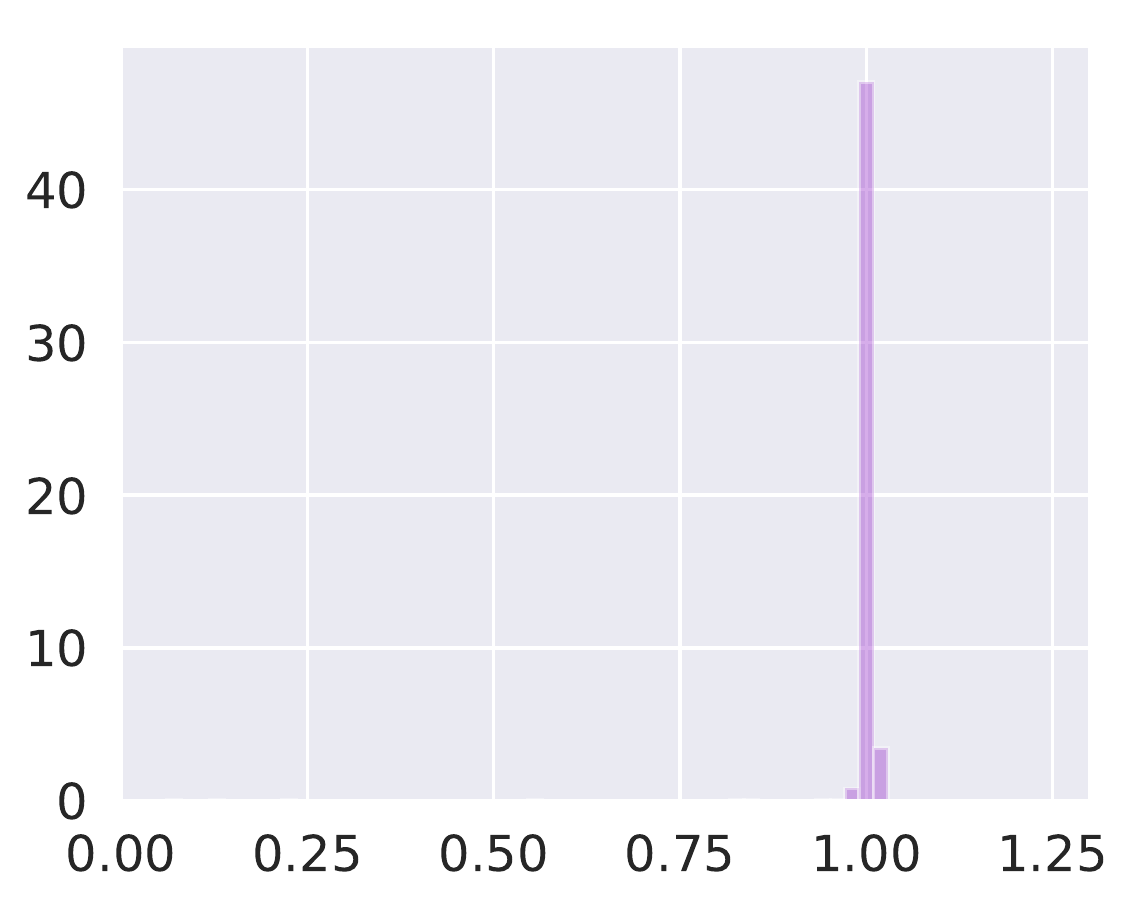}\nsp &
  \nsp\includegraphics[height=\f1ht]{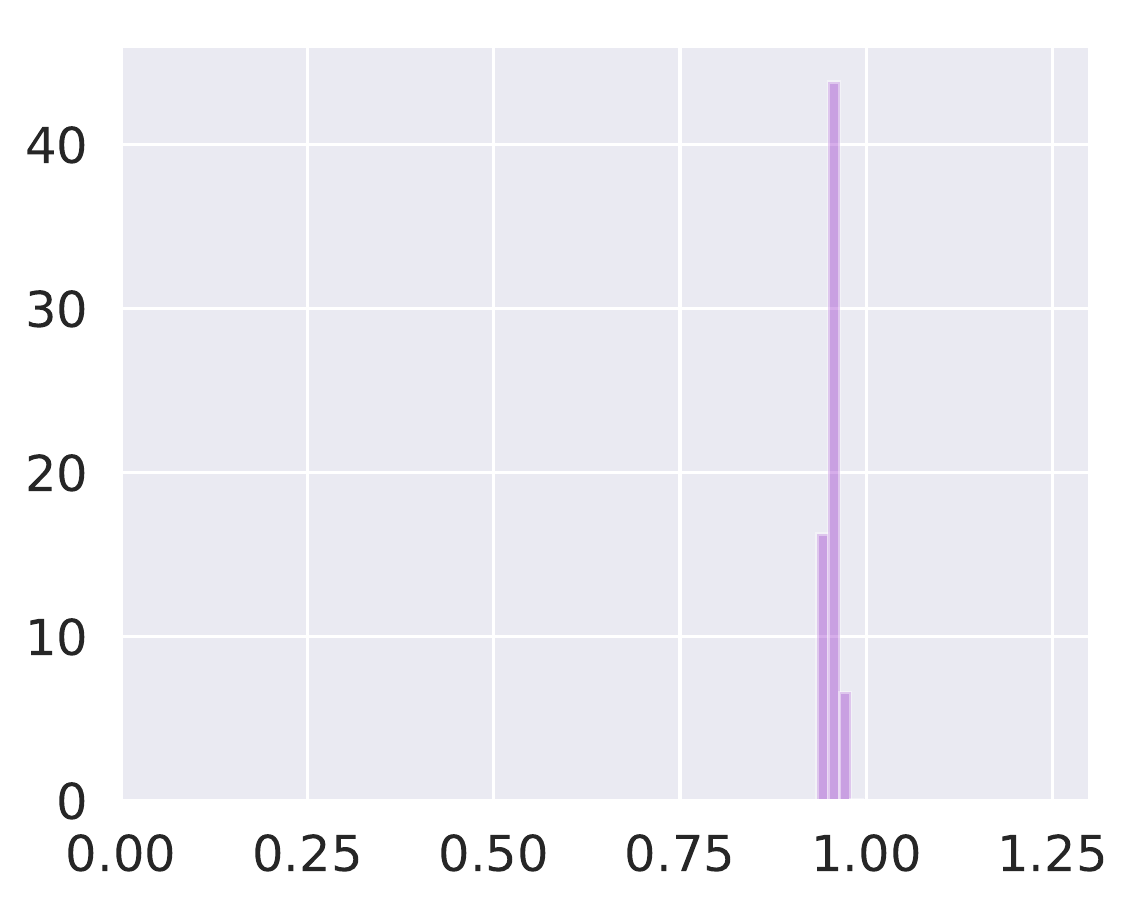}\nsp &
  \nsp\includegraphics[height=\f1ht]{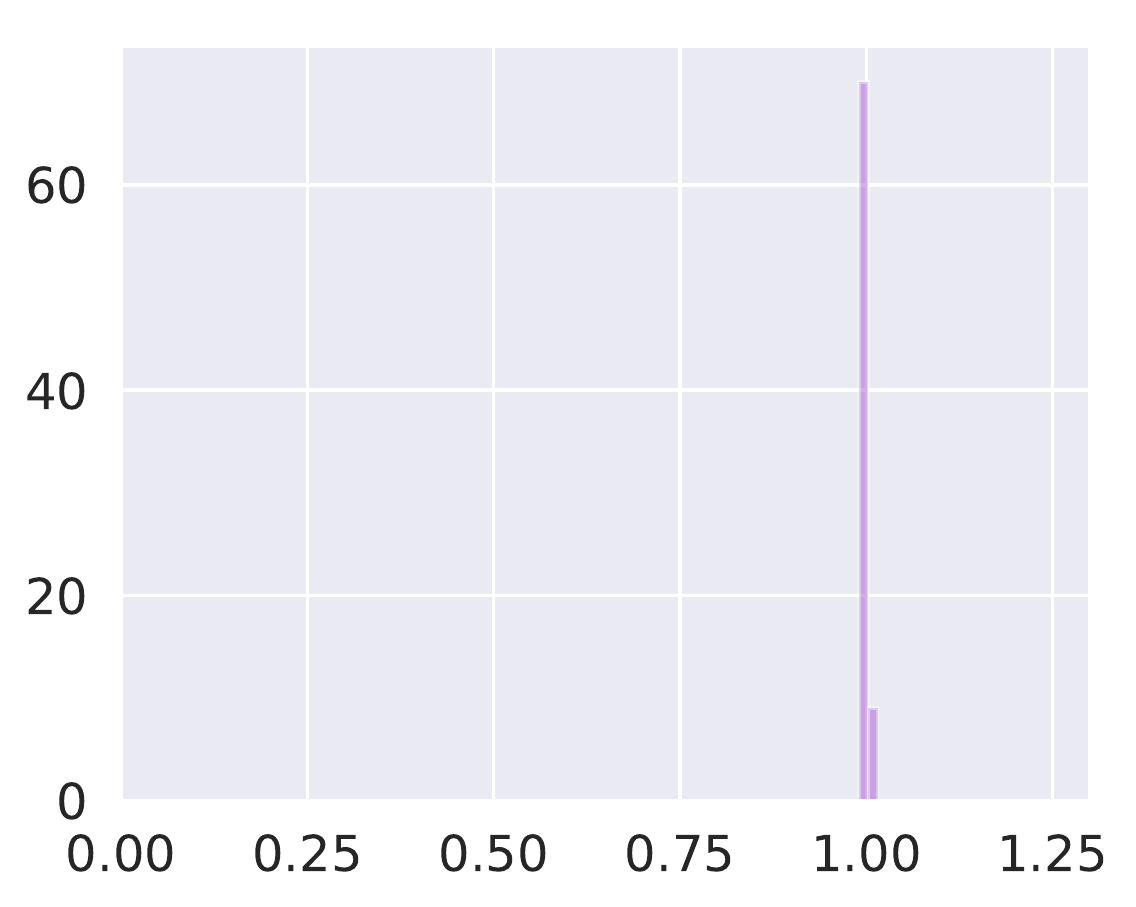} \\
  \end{tabular}\\[-0.5ex]
\caption{Histograms of the standard deviation of normalized intermediate features each corresponding to a separate convolutional filter for the CIFAR-10-C, JIGSAWS, and StrokeRehab datasets. On the top row, the normalization is carried out on test data using statistics estimated from the training data (i.e. non-adaptively). The histogram of standard deviations does not concentrate around one, as opposed to the standard-deviation histogram in Figure~\ref{fig:bn_cifar10_fmnist}. This indicates that the non-adaptive normalization scheme is not successful in normalizing the held-out data. On the bottom row, the normalization is carried out adaptively, using batches where the value of the extraneous variable is fixed. This results in a histogram that does concentrate around one, demonstrating that adaptive normalization succeeds in normalizing the features correctly.}
\label{fig:hist_std_ext}
\end{figure*}

In many applications, the observed data depend heavily on \emph{extraneous} variables that are not directly relevant to the task of interest. Changes in these variables produce significant shifts in the distribution of the data. If a machine-learning model is trained with data corresponding to a certain value of an extraneous variable, its performance may be significantly degraded when tested on data corresponding to a different value. Unfortunately, this is a common occurrence in practical applications when machine-learning models often need to be applied in changing environments. In this work, we consider the problem of performing classification in the presence of extraneous variables with fluctuating values. As in many real-world scenarios, the values of the extraneous variables observed during inference \emph{are not available during training}. This is in stark contrast to most existing literature on domain adaptation, where it is usually assumed that unlabeled data corresponding to the target domain are available during training~\cite{ganin2016domain, tzeng2017adversarial,long2013transfer, long2017deep, long2016unsupervised}.

In this paper we study three different datasets illustrating realistic situations where it is crucial to generalize across unknown values of the extraneous variables:
\begin{itemize}[topsep=0pt,itemsep=-0.5ex,partopsep=1ex,parsep=2ex]
    \item The JIGSAWS dataset, where the goal is to classify the different gestures performed by surgeons during surgery from sensor data. Here an important extraneous variable is the identity of the surgeon; different surgeons will move in systematically different ways during surgery. The ultimate goal of this task is to aid in surgeon training, so it is imperative for models to generalize to new surgeons.
    \item The StrokeRehab dataset, where the goal is to classify upper-body movements in stroke patients from sensor data. Similarly to the previous case, the identity of the patient is an important extraneous variable, since different patients have different levels of impairment, as well as other idiosyncrasies. Generalization to new patients during inference is critical if these models are to be deployed in realistic rehabilitation settings.
    \item The CIFAR-10C dataset, where the goal is to classify natural images that are subject to common corruptions such as blurring, compression artifacts, additive noise, etc. Here the extraneous variable is the type of corruption. Making computer vision models robust to unexpected corruptions is crucial in real-world deployment, e.g. for self-driving vehicles.
\end{itemize}
The main contributions of this work are the following. We identify a crucial weakness of batch normalization in settings where extraneous variables vary between training and inference: normalization is carried out with inexact estimates of the feature statistics. To alleviate this issue, we propose estimating these features adaptively at test time, by averaging either over batches with a fixed value of the extraneous variable, or over single data instances. We then demonstrate that adaptive normalization results in gains of between 10\% and 15\% in classification performance, with respect to non-adaptive schemes, for different network architectures and choices of feature statistics. Finally, we evaluate the invariance of the learned features to the extraneous variables, showing that in some cases adaptive normalization results in increased invariance. 


\section{Related Work}
\label{sec:related_work}

Traditional approaches to the control of extraneous variables in statistics often rely on experimental design and randomization~\cite{box2005statistics}, and hence are not applicable to observational studies. Here we restrict the discussion to techniques that are applicable to observational data, although some are also applied in randomized settings.

Frequentist techniques to perform inference in the presence of extraneous variables are often parametric; they assume that we have available the likelihood of the data as a function of the response (i.e. the variable that we want to estimate) and the extraneous variables. In that case, one can perform inference by maximizing the so-called profile likelihood, obtained by first determining the maximum likelihood estimate of the extraneous variables given the response, and then plugging the estimate into the likelihood function
~\cite{reid2003likelihood,barndorff1983formula,cox1987parameter,mccullagh1990simple}. An alternative to profile-likelihood methods is based on applying transformations that decouple the effects of the response and the extraneous variables on the data. More concretely, the goal is to find a sufficient statistic whose marginal or conditional distribution depends on the response, but not on the extraneous variables. The distributions can then be used as a likelihood function for the response. These approaches are known as marginalization and conditioning in the literature~\cite{basu1977elimination}. From a Bayesian point of view, inference with respect to the response can be carried out after marginalizing with respect to the extraneous variable. This requires endowing the extraneous variables with a prior. When noninformative priors are used, this is sometimes denoted as an integrated likelihood approach~\cite{berger1999integrated}. 

Unfortunately, when the data are high dimensional, it is usually not tractable to derive a realistic likelihood function that accurately captures the interactions between the extraneous variables and the response, except in very restricted scenarios. Nonparametric approaches to the control of extraneous variables often rely on methodology derived under linearity assumptions, such as analysis of variance~\cite{neter1996applied} or variable-selection techniques~\cite{guyon2003introduction}. A recent method based on optimal transport~\cite{tabak2018explanation} learns a linear map that transforms the data to make it as invariant as possible to the effect of extraneous variables. 

In the last 15 years, there has been significant interest in removing batch effects from high-throughput genomic data has been an active area of research. Batch effects are due to extraneous variables dependent on experimental
variations, such as intensity changes in the scanning process, block effects, dye effects, or amount of mRNA and DNA concentration on microarrays~\cite{lazar2012batch}. There are two main approaches for batch-effect removal. Location-scale methods normalize the data within the batches by centering~\cite{sims2008removal} and normalizing it~\cite{li2001model,kim2007attempt}. Matrix-factorization methods assume that variation due to batch effects is independent of the variation due to biological variables of interest, and can be captured by a small set of factors estimated by fitting a low-rank factor model to the data~\cite{leek2007capturing}, for instance via the singular-value decomposition~\cite{alter2000singular}.

At a high level, most current statistical techniques for the control of extraneous variables apply transformations based on probabilistic, linear or low-rank assumptions in order to normalize the data and remove their extraneous effects. Unfortunately, these techniques do not easily translate to high-dimensional problems requiring nonlinear feature extraction, because these assumptions do not hold. Instead, in this work we take a different route, showing that adaptive feature normalization provides some robustness to changes in extraneous variables. Adaptive normalization has been described as a canonical neural computation in neuroscience because biological systems employ it on essentially all sensory modalities~\cite{carandini2012normalization}. In machine learning it has been applied to natural image compression~\cite{balle2016end}, density estimation~\cite{balle2015density}, style transfer~\cite{ulyanov2016instance}, and image classification~\cite{singh2019filter}.  


\section{Batch Normalization} \label{sec:batch_norm}



Batch normalization is a technique that increases the robustness of deep-learning models to changes in initialization and learning rates~\cite{ioffe2015batch}. Due to its widespread success, it has become a standard component of deep convolutional neural networks. It consists of two steps, typically applied at each layer of the neural network: (1) centering and normalizing the intermediate features corresponding to each convolutional filter using an estimate of their mean and standard deviation, (2) scaling and shifting the normalized features using two additional learned parameters per filter (a scaling factor and a shift). During training, the mean and standard deviation are computed by averaging the features over each training batch. In addition, a running average of the mean and the variance are used to obtain a global estimate of the population mean and standard deviation of each feature. These global estimates are used to perform normalization during inference.

Understanding why batch normalization aids in the training of deep neural networks is an active area of research. Originally, it was hypothesized that it reduces internal covariate shift~\citep{ioffe2015batch}; more recent work suggests that it smooths the optimization landscape of the training cost function~\citep{santurkar2018does}. Regardless of the precise mechanisms associated with it, batch normalization operates under a basic assumption: the mean and standard deviation of the network features are \emph{the same} during training and inference. This assumption holds for many standard benchmarks where the training and test data follow the same distribution, such as CIFAR-10~\citep{krizhevsky2009learning} and Fashion-MNIST~\citep{xiao2017online}. To demonstrate this, we use feature statistics estimated from the training set to center and normalize the corresponding features computed from test data. Figure~\ref{fig:bn_cifar10_fmnist} shows the histograms of the means and standard deviations from the centered, normalized features. As expected, they concentrate around zero and one respectively. 

For data that depends on extraneous variables, which change at test time, the training statistics may no longer resemble the train statistics. To evaluate this, we perform the same experiment as in Figure~\ref{fig:bn_cifar10_fmnist} for our three datasets of interest (see Section~\ref{sec:datasets} for a detailed description). In this case, many of the centered, normalized test features have means and standard deviations that deviate significantly from zero and one respectively (Figures~\ref{fig:hist_mean_ext}, \ref{fig:hist_std_ext}).

\section{Adaptive Normalization} \label{sec:adaptive_normalization}

As reported in the previous section, changes in extraneous variables may significantly affect the mean and standard deviation of filters in convolutional neural networks. As we report in Section~\ref{sec:results}, this mismatch is accompanied by poor classification performance. A natural solution for this issue is to perform normalization \emph{adaptively} during inference, using statistics estimated from the test data. When the value of the extraneous variable is known, then this can be achieved by feeding a batch of test examples corresponding to a fixed value of the extraneous variable through the network, and averaging the resulting features to estimate the means and standard deviations for each filter. To evaluate whether this approach results in better normalization, we divide the test set of our three datasets of interest into subsets associated with different values of the relevant extraneous variable (surgeons for JIGSAWS, patients for StrokeRehab, and corruptions for CIFAR-10C). For each subset, we estimate the feature statistics from one-half of the data and use the corresponding statistics to center and normalize the other half. Figures~\ref{fig:hist_mean_ext}, \ref{fig:hist_std_ext} show that adaptive normalization fulfils its goal: the means and standard deviations are concentrated around zero and one respectively.

Extraneous variables may be easy to identify during inference for some datasets: for example, when the extraneous variable is associated with different subjects, as in JIGSAWS (surgeons) or StrokeRehab (stroke patients), it is reasonable to assume that this information may be known at test time. However, extraneous variables may also be unknown, or not easy to identify. In such situations, we can still apply adaptive normalization by estimating the feature statistics \emph{from each test instance}. This ensures that the statistics are performed for a fixed value of the extraneous variables. However, averaging only over one instance results in a high variance in the estimated statistics, resulting in a mismatch with the statistics used to perform feature normalization during training. This suggests an additional modification: using instance-based averaging also during training. The resulting normalization scheme is known as instance normalization. It was originally introduced in the context of style transfer to promote invariance to style variations in images~\cite{ulyanov2016instance}. 

Adaptive normalization ensures that features at each layer of a convolutional neural network are normalized similarly at training and inference, even when extraneous variables associated with the data vary. Intuitively, this is a desirable property, because it prevents spurious shifts and scalings that may distort the features, and ultimately degrade classification accuracy. In Section~\ref{sec:results}, we provide numerical evidence to support this hypothesis: both batch-based and instance-based adaptive normalization significantly improve the classification accuracy for our datasets of interest. Motivated by recent work~\citep{singh2019filter}, we also compare non-adaptive and adaptive normalization based on mean square (as opposed to mean and standard deviation). The gains in accuracy achieved by adaptive normalization are similar. 

Our observations are consistent with recent work by \cite{pan2018two}, who propose to apply batch-based non-adaptive normalization and instance-based adaptive normalization to different filters in a convolutional network. They show that this results in improved performance on a domain-adaptation task, without fine-tuning to the target domain. Our results suggest that this generalization ability may be mostly due to the use of adaptive normalization. 


\section{Datasets}
\label{sec:datasets}

\begingroup
\renewcommand*{\arraystretch}{1.5}
\begin{table*}
\centering
\begin{tabular}{|c|c|c|c|c|c|c|}
\hline
Normalization scheme & \multicolumn{2}{c|}{Non-adaptive} & \multicolumn{4}{c|}{Adaptive}                                          \\ \hline
Averaging scheme            & \multicolumn{2}{c|}{Batch based}  & \multicolumn{2}{c|}{Instance based} & \multicolumn{2}{c|}{Batch based} \\ \hline
Statistics           & $\mu,\sigma$      & $\nu^2$     & $\mu,\sigma$    & $\nu^2$         & $\mu,\sigma$     & $\nu^2$     \\ \hline
JIGSAWS              & 67.22               & 64.88       & \textbf{72.71}              & 72.51   & 65.51              & 62.55       \\
StrokeRehab          & 48.3                & 44.91       & \textbf{63.9}     & 63.8            & 61.6               & 60.6        \\ 
CIFAR-10C            & 72.57               & 70.75       & 81.58             & 80.52           & \textbf{81.88}     & 78.14       \\ \hline
CIFAR                & \textbf{94.15}      & 91.27       & 92.08             & 90.99           & 91.48              & 87.47       \\
Fashion-MNIST         & \textbf{92.38}      & 91.29       & 90.52             & 90.22           & 88.55              & 90.68       \\ \hline
\end{tabular}

\caption{Results of the computational experiments described in Section~\ref{sec:experiments}. The entries in the table report the classification performance for different normalization strategies: non-adaptive or adaptive, instance-based or batch-based, and different statistics. For the statistics, $\mu$ denotes the mean, $\sigma$ denotes the standard deviation, and $
\nu^2$ denotes the mean square. For CIFAR-10 and Fashion-MNIST, the non-adaptive batch-based normalization schemes outperform all others, by small margins. In sharp contrast, for the datasets dependent on extraneous variables, adaptive normalization schemes outperform non-adaptive schemes by around 10\% in JIGSAWS, 15\% in StrokeRehab and 10\% in CIFAR-10C. Similar results for other test-train splits of JIGSAW dataset is given in table \ref{tab:jigsaws_multi_split}.}
\label{tab:accuracy_normalization}
\end{table*}
\endgroup

In this section, we describe the datasets used in our computational experiments. We use two standard benchmarks for classification algorithms to represent situations where the data do not depend on extraneous variables: CIFAR-10, a dataset of $32 \times 32$ natural color images belonging to $10$ different classes \cite{krizhevsky2009learning}, and  
Fashion-MNIST, a dataset of $28 \times 28$ grayscale images of 10 different fashion items~\citep{xiao2017online}. These datasets are compared to three datasets, which are influenced by known extraneous variables. These datasets, described in detail below, are the main focus of our experiments.

\subsection{Surgical activities}
\label{sec:jigsaws}
The JHU-ISI Gesture and Skill Assessment Working Set (JIGSAWS) dataset introduced in \citep{gao2014jhu} contains data measuring the motion of eight different surgeons with different skill levels performing repetitions of three basic surgical tasks (suturing, knot-typing and needle-passing) on a benchtop model. The original dataset contains both video and kinematic sensor data. In our experiments, we only use the kinematic data, which contains cartesian positions, orientations, velocities, angular velocities and gripper angle describing the motion of the robotic manipulators used to perform the surgical tasks. These data correspond to a $76$-dimensional time series. Each time step is manually annotated with a label indicating one of 15 possible gestures or atomic surgical activities carried out by the surgeon at that time. The classification task that we consider is classifying these gestures automatically from the sensor data. The main extraneous variable influencing the data is the identity of the surgeon because different people move in distinctive ways while performing surgery.


\subsection{Stroke rehabilitation}
\label{sec:sensor}

The StrokeRehab dataset contains motion data collected while 32 stroke patients performed a battery of activities of daily living (including feeding, drinking, washing face, brushing teeth, donning glasses, etc.). The data were recorded with inertial measurement units (IMUs) affixed to the upper body of the patients. Each IMU sensor samples 3D linear acceleration and 3D angular velocity at 100 Hz. Data are integrated and fused algorithmically to provide acceleration, rotation, and quaternion values for each 3D axis. The data consequently corresponds to a 78-dimensional time series. Each multidimensional time sample in the dataset is labeled with one of five \emph{movement primitives} that the patient was performing at the time (reach, idle, transport, stabilize and reposition). Movement primitives are discrete motions that can be strung together in various combinations to build most basic arm motions. In this dataset, a crucial extraneous variable is the identity of the patient. Each patient is partially paralyzed with varying levels of impairment. This results in highly idiosyncratic movement patterns associated with each patient.

\subsection{Corrupted natural images}
\label{sec:cifar}

The CIFAR-10C Benchmark Dataset\footnote{Available for download at \url{https://github.com/hendrycks/robustness}} is a standardized dataset designed to evaluate the robustness of image classifiers to different natural-image corruptions\citep{hendrycks2019robustness}. The data consist of the $10,000$ validation images in CIFAR-10 perturbed by 19 different corruptions: brightness, contrast, defocus blur, elastic transform, fog, frost, gaussian blur, gaussian noise, glass blur, impulse noise, jpeg compression, motion blur, pixelate, saturate, shot noise, snow, spatter, speckle noise and zoom blur. Each corruption is applied at $5$ different severity levels, for a total of $50,000$ images per corruption class. The task is to classify the corrupted images into 10 different classes using models trained on uncorrupted images. The type of corruption is an extraneous variable because it produces systematic differences in the images.






\section{Computational Experiments}
\label{sec:experiments}

\begingroup
\renewcommand*{\arraystretch}{1.5}
\begin{table*}
\centering
\begin{tabular}{|c|c|c|c|c|c|c|}
\hline
Normalization scheme & \multicolumn{2}{c|}{Non-adaptive} & \multicolumn{4}{c|}{Adaptive}                                          \\ \hline
Averaging scheme & \multicolumn{2}{c|}{Batch based}  & \multicolumn{2}{c|}{Instance based} & \multicolumn{2}{c|}{Batch based} \\ \hline
Statistics           & $\mu,\sigma$      & $\nu^2$     & $\mu,\sigma$    & $\nu^2$         & $\mu,\sigma$     & $\nu^2$     \\ \hline
ResNet50              & 21.58               & 20.69       & 10.5              & 10.47   & 9.6              & \textbf{9.33}       \\
VGG19          & 20.82                & 21.27       & 12.06     & 14.07            & \textbf{9.04}               & 11.61        \\ 
DenseNet121                & 23.05      & 24.79       & 9.36             & 11.1           & \textbf{8.81}              &  10.74       \\
\hline
\end{tabular}

\caption{Results of the computational experiments described in Section~\ref{sec:experiments} on the CIFAR-10C dataset for several different architectures. The entries in the table report the drop in classification performance of the networks relative to their performance on CIFAR-10. The lower the drop in accuracy, the better the relative performance of the network. Results are reported for different normalization strategies: non-adaptive or adaptive, instance-based or batch-based, and different statistics. For the statistics, $\mu$ denotes the mean, $\sigma$ denotes the standard deviation, and $\nu^2$ denotes the mean square. For all architectures, adaptive normalization achieves around 10\% improvement over non-adaptive normalization. }
\label{tab:accuracy_cifar10c}
\end{table*}
\endgroup

In this section, we describe our computational experiments to study the effect of different normalization schemes on the performance of deep convolutional neural networks trained for classification. Our experiments involve two types of datasets: \emph{homogeneous} datasets where there is no variation of extraneous variables between the training and test sets (CIFAR-10 and Fashion-MNIST), and \emph{heterogeneous} datasets where the data distribution is influenced by extraneous variables that do change between training and test (JIGSAWS, StrokeRehab, and CIFAR-10C). For CIFAR-10 and Fashion-MNIST, we use the usual training and test sets. For the heterogeneous datasets, the data are randomly separated according to the different extraneous variables in the following way:
\begin{itemize}[topsep=0pt,itemsep=-0.5ex,partopsep=1ex,parsep=2ex]
        \item JIGSAWS: The training set consists of data from 8 surgeons. The validation and test sets consist of data from 2 different surgeons.
        \item StrokeRehab: The training set consists of data from 21 patients. The validation and test sets consist of data from 5 and 6 different patients respectively.
        \item CIFAR-10C: The training and validation data sets are the same as in CIFAR-10. The test set consists of 19 different corrupted versions of the CIFAR-10 validation set.
    \end{itemize}

For each dataset, we train deep convolutional networks with different feature-normalization strategies that are combinations of three basic elements:
\begin{enumerate}[topsep=0pt,itemsep=-0.5ex,partopsep=1ex,parsep=2ex]
    \item \textbf{Normalization scheme}, which can be \emph{non-adaptive}-- as in traditional batch normalization where the statistics are estimated during training and then fixed-- or \emph{adaptive}-- as in instance normalization or feature response normalization where the statistics are re-estimated at test time.
    \item \textbf{Averaging scheme}, which can be \emph{instance-based}-- as in instance normalization or feature response normalization where the filter responses are averaged over a single data point-- or \emph{batch-based}-- as in batch normalization where batches with several data points are used.
    \item \textbf{Statistics}, which are either the mean and standard deviation-- as in batch and instance normalization-- or the mean square-- as in feature response normalization.
\end{enumerate}
We apply all possible combinations of these elements, except for instance-based averaging and non-adaptive normalization because non-adaptive normalization requires stable estimates of the statistics.  

The neural networks used in our experiments are based on standard convolutional architectures for classification. For the JIGSAWS and StrokeRehab datasets, we use a variation of DenseNet~\citep{huang2017densely} with 11 dense blocks each consisting of four sets of convolutions, ReLU nonlinearity and a normalization layers. For CIFAR-10, CIFAR-10C and Fashion-MNIST, we use VGG-19~\citep{simonyan2014very}, Resnet-50~\citep{he2016deep} and DenseNet-121~\citep{huang2017densely} architectures. We provide additional details about the architectures in section \ref{app:sec:arch_design_train}. All these architectures are modified to apply different normalization strategies to the intermediate features.

For the sensor datasets (JIGSAWS and StrokeRehab) we perform classification on windows of 200 time steps (acquisition frequency 30 Hz for JIGSAWS and 100 Hz for StrokeRehab). We use overlapping windows with a stride of 25 time steps for training. The windows in the validation and test sets are non-overlapping. To remove spurious offsets caused by sensor calibration, we center and normalize the sensor data per dimension. For CIFAR-10 and CIFAR-10C, we center and normalize the RGB channels using channel mean and standard deviations computed from the CIFAR-10 training set. We use the Adam optimizer~\cite{kingma2014adam} to train all models varying the learning rate when the validation accuracy plateaus (see section \ref{app:sec:arch_design_train}). We apply early stopping based on the validation set.

\begingroup
\renewcommand*{\arraystretch}{1.5}
\begin{table*}[t]
\centering
\begin{tabular}{|c|c|c|c|c|c|c|c|}
\hline
Normalization scheme & \multicolumn{2}{c|}{Non-adaptive} & \multicolumn{4}{c|}{Adaptive}                                          & {\begin{tabular}[c]{@{}c@{}}Raw Signal\\\end{tabular}} \\ \cline{1-7}
\textbf{}            Averaging scheme& \multicolumn{2}{c|}{Batch based}  & \multicolumn{2}{c|}{Instance based} & \multicolumn{2}{c|}{Batch based} &                                                                                            \\ \cline{1-7}
Statistics           & $\mu,\sigma$      & $\nu^2$     & $\mu,\sigma$       & $\nu^2$      & $\mu,\sigma$     & $\nu^2$     &                                                                                            \\ \hline
JIGSAWS              & 94.53               & 95.80       & 51.15       & 61.19        & 91.41              & 86.5        & 98.4                                                                               \\ \hline
StrokeRehab          & 24.60      & 31.34       & 26.0                 & 28.26         & 48.53              & 50.44       & 68.0                                                                              \\ \hline
CIFAR-10C            & 50.57               & 54.97       & 21.92                & 30.28        & 16.93    & 24.31       & 78.03                                                                              \\ \hline
\end{tabular}
\caption{Accuracy of decoding the value of the relevant extraneous variables from the features in the penultimate layer of convolutional networks trained using different normalization schemes. Networks trained with adaptive normalization schemes tend to have lower decoding accuracy as compared to the ones trained with non-adaptive normalization scheme for the CIFAR-10C and JIGSAWS datsets, which suggests higher invariance to the variations in the extraneous variable. For comparison, the accuracy of randomly guessing the value of the extraneous variable equals 12.5\% for JIGSAWS, 3.125\% for StrokeRehab, and 5.26\% for CIFAR-10C.}
\label{tab:decoding_extraneous_var}
\end{table*}
\endgroup



\section{Results}
\label{sec:results}




\begin{figure}[t]
\centering
\includegraphics[width=\linewidth]{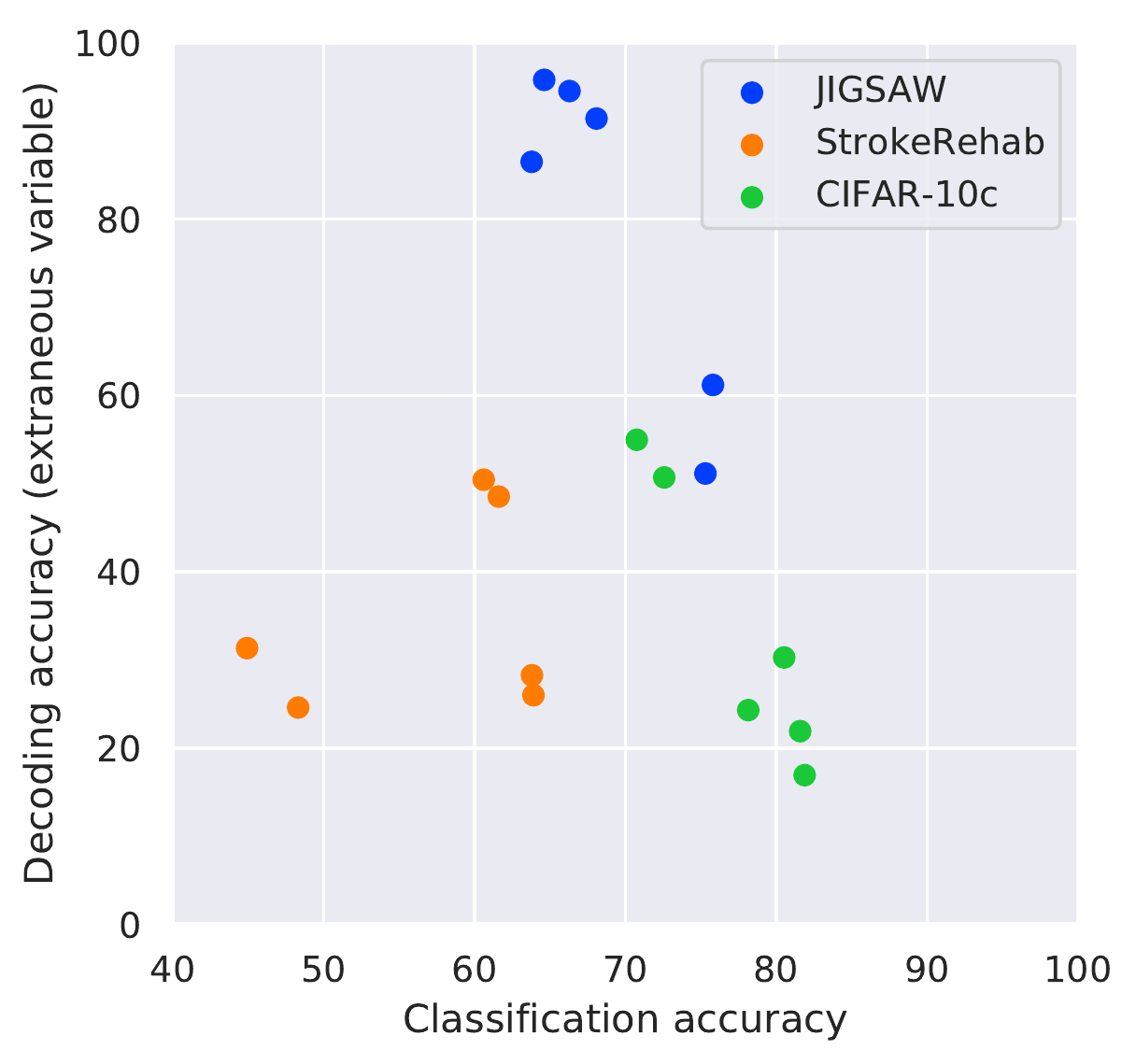}
\caption{Scatterplot of classification accuracy and accuracy of decoding the value of the relevant extraneous variable (as a proxy for invariance). For each dataset, the classification accuracy is higher the lower the decoding accuracy, with the exception of the StrokeRehab dataset, where the features normalized non-adaptively result in very poor classification and decoding accuracy.}
\label{fig:class_vs_decoding}
\end{figure}

Table~\ref{tab:accuracy_normalization} reports the results of the computational experiments described in Section~\ref{sec:experiments}. For CIFAR-10 and Fashion-MNIST, the non-adaptive batch-based normalization schemes outperform all others, by small margins. In sharp contrast, for the datasets dependent on extraneous variables, adaptive normalization schemes outperform non-adaptive schemes by around 10\% in JIGSAWS, 15\% in StrokeRehab and 10\% in CIFAR-10C. Instance-based adaptive normalization performs around 7\% better than batch-based adaptive normalization in JIGSAWS, whereas both adaptive schemes perform similarly in StrokeRehab and CIFAR-10C. This is consistent with Figures~\ref{fig:hist_mean_ext} and~\ref{fig:hist_std_ext}, where batch-based adaptive normalization is not as successful for JIGSAWS as for the two other datasets. This could indicate the presence of other extraneous variables, apart from the surgeon identity. Finally, normalization based on mean and standard deviation, and on mean square perform similarly in most cases. 

Table~\ref{tab:accuracy_cifar10c} shows the accuracy of the different normalization schemes on CIFAR-10C when compared to CIFAR-10 for different architectures (Table~\ref{tab:accuracy_normalization} only includes the best architecture in each case). The conclusions are essentially the same: for all architectures, adaptive normalization achieves around 10\% improvement over non-adaptive normalization. Batch-based adaptive normalization slightly outperforms instance-based normalization. Normalization based on mean and standard deviation, and on mean square again perform similarly.

The outcomes of our experiments confirm our hypothesis for our datasets of interest: correcting the mismatch in feature statistics produced by variations in extraneous variables systematically results in significantly higher classification accuracy. 

\section{Invariance Analysis}
\label{sec:invariance_analysis}

The ultimate goal when designing machine-learning methods that are robust to variations in extraneous variables is to learn features that are (1) useful for the task of interest, and (2) as invariant as possible with respect to these variables. In this section, we study to what extent the features learned by convolutional neural networks with different normalization schemes are invariant to extraneous variables for our three datasets of interest. Our strategy is to train small neural networks to decode the value of the extraneous variable from the features at the penultimate layer of the networks (further details about these networks are provided in section \ref{app:sec:train_details_invariance_analysis}). Invariance is then quantified by evaluating the decoding accuracy achieved from features corresponding to held-out data. The lower this accuracy, the more invariant the network is to the extraneous variable. 

Table~\ref{tab:decoding_extraneous_var} shows the results. In general, models trained with adaptive normalization seem to be more invariant to the extraneous features. Figure~\ref{fig:class_vs_decoding} shows a scatterplot of classification accuracy and decoding accuracy (as a proxy for invariance). For each dataset, the classification accuracy is higher the lower the decoding accuracy, with the exception of the StrokeRehab dataset, where the features normalized non-adaptively result in very poor classification and decoding accuracy. It is worth emphasizing that adaptive normalization is designed to only ensure approximate invariance of the first and second-order statistics of the features. Our results suggest that the resulting features tend to be more invariant to extraneous variables for the JIGSAW and CIFAR-10C datasets. It is worth noting that several works~\cite{ulyanov2016instance,huang2017arbitrary} have provided evidence that instance-based adaptive normalization boosts invariance to changes in image style, in the context of style transfer.

\section{Conclusion}
Designing robust methodology capable of performing successfully in changing environments is a fundamental goal in modern machine learning. In this work, we show that when processing data that is heavily dependent on extraneous variables, one should-- in the words of Bruce Lee-- \emph{be like water} and adapt to the changing environment. More concretely, we demonstrate that standard non-adaptive feature normalization fails to correctly normalize the features of convolutional neural networks on held-out data where extraneous variables take values not seen during training. This can be remedied by normalizing adaptively, with statistics computed from the held-out data itself. In experiments with three different datasets affected by extraneous variables, switching to adaptive normalization consistently results in significant increases in classification accuracy. In addition, for two of the datasets, adaptive normalization seems to produce an increased invariance to the extraneous variable (in the sense that it becomes more challenging to decode the variable from the learned features). A challenge for future research is to further boost invariance to extraneous variables, even when the values of those variables are not known beforehand. 

\section*{Acknowledgements}
We would like to thank Prof. Kyunghyun Cho for useful discussions. We would also like to thank the volunteers who contributed to label the StrokeRehab dataset: Ronak Trivedi, Adisa Velovic, Sanya Rastogi, Candace Cameron, Sirajul Islam, Bria Bartsch, and Courtney Nilson. This research was supported by NIH grants R01 LM013316 (AK, CFG, HMS) and K02 NS104207 (HMS).



\nocite{langley00}

\bibliography{example_paper}
\bibliographystyle{icml2019}

\appendix

\section{Additional results}
\subsection{JIGSAWS}
In this section we report additional results for the surgical-activity dataset (JIGSAWS). The extraneous variable of interest for these data is the identity of the surgeon. There are eight surgeons in the dataset. We report the results of training on six, using one for validation, and then testing on the remaining one for six different random splits. The result for all the splits are reported in Table~\ref{tab:jigsaws_multi_split}. They confirm the result reported in the main paper (which is the mean of all the splits): instance-based adaptive normalization consistently outperforms all other normalization schemes for these data.

\begingroup
\renewcommand*{\arraystretch}{1.5}
\begin{table*}[t]
\centering
\begin{tabular}{|c|c|c|c|c|c|c|}
\hline
Normalization scheme & \multicolumn{2}{c|}{Non-adaptive} & \multicolumn{4}{c|}{Adaptive}                                          \\ \hline
Averaging scheme            & \multicolumn{2}{c|}{Batch based}  & \multicolumn{2}{c|}{Instance based} & \multicolumn{2}{c|}{Batch based} \\ \hline
Statistics           & $\mu,\sigma$      & $\nu^2$     & $\mu,\sigma$    & $\nu^2$         & $\mu,\sigma$     & $\nu^2$     \\ \hline
Split - 1 & 66.29               & 64.61       & 75.3              & \textbf{75.8}   & 68.08              & 63.78 \\
Split - 2             & 56.91               & 54.63 & \textbf{58.64}              & 58.39   &  54.08             &  46.42      \\
Split - 3         & 70.79               & 66.19       & 73.6     & \textbf{76.64}            & 65.31               & 64.33         \\ 
Split - 4            &      67.93          & 67.79       & 76.15             & \textbf{76.49}           &  67.93     & 64.16        \\ 
Split - 5             & 67.59      & 64.98       & \textbf{72.21}             &  70.15          & 70.16             &  70.42       \\
Split - 6        & 73.82      & 71.41       & \textbf{80.38}            & 77.61           & 67.55              & 66.24       \\ \hline
\end{tabular}

\caption{Results of the computational experiments described in section \ref{sec:experiments} on different splits of the JIGSAWS dataset. The entries in the table indicate the classification performance for different normalization strategies: non-adaptive or adaptive, instance-based or batch-based, and different statistics. For the statistics, $\mu$ denotes the mean, $\sigma$ denotes the standard deviation, and $
\nu^2$ denotes the mean square. For all splits, instance-based adaptive normalization outperforms all other normalization schemes.}
\label{tab:jigsaws_multi_split}
\end{table*}
\endgroup
    
\subsection{Invariance Analysis}
In section \ref{sec:invariance_analysis}, we analyze to what extent features learned by different models are invariant to  extraneous variables, by trying to decode the extraneous variables from the features. In this section, we provide additional results evaluating invariance in different layers of the networks. Tables~\ref{tab:jigsaws_decoding_acc_layers}, \ref{tab:strokerehab_decoding_acc_layers} and \ref{tab:cifar10c_decoding_acc_layers} report the results. We observe two patterns: 1) networks trained with instance-based adaptive normalization learn features that are more invariant to extraneous variables 2) the features from deeper layers of the network are more invariant to extraneous variables.

\begingroup
\renewcommand*{\arraystretch}{1.5}
\begin{table*}[t]
\centering
\begin{tabular}{|c|c|c|c|c|c|c|}
\hline
Normalization scheme & \multicolumn{2}{c|}{Non-adaptive} & \multicolumn{4}{c|}{Adaptive}                                           \\ \cline{1-7}
\textbf{}            Averaging scheme& \multicolumn{2}{c|}{Batch based}  & \multicolumn{2}{c|}{Instance based} & \multicolumn{2}{c|}{Batch based}                                                                                            \\ \cline{1-7}
Statistics           & $\mu,\sigma$      & $\nu^2$     & $\mu,\sigma$       & $\nu^2$      & $\mu,\sigma$     & $\nu^2$                                                                                            \\ \hline
Layer - 24             & 97.54               & 99.00      & 80.90       & 83.02        & 96.69              & 95.00     \\ 
Layer - 28          & 96.22      & 97.40       & 68.22   & 77.93         & 93.77              & 93.25       \\
Layer - 32            & 95.47               & 96.88       & 65.58                & 80.95        & 92.36    & 91.65         \\
Layer - 36            & 96.27               & 97.21       & 61.99                & 69.54        & 92.03    & 89.53        \\ 
Layer - 40            & 95.00               & 96.88       & 57.37                & 58.93        & 83.21    & 80.66             \\
Layer - 44 (Penultimate Layer)           & 94.53               & 95.80       & 51.15       & 61.19        & 91.41              & 86.5     \\ \hline
\end{tabular}

\caption{Accuracy of decoding the identity of each surgeon in the \textbf{JIGSAWS} dataset from features at different layers of convolutional networks trained using different normalization schemes. Networks trained with adaptive normalization schemes tend to have lower decoding accuracy than the ones trained with non-adaptive normalization scheme, which suggests higher invariance to the extraneous variable. Instance-based normalization provides significantly more invariance (and also better performance as shown in Table~\ref{tab:jigsaws_multi_split}). Additionally, features from deeper layers show more invariance. For comparison, the accuracy of randomly guessing the value of the extraneous variable equals 12.5\% for this dataset.}
\label{tab:jigsaws_decoding_acc_layers}
\end{table*}
\endgroup

\begingroup
\renewcommand*{\arraystretch}{1.5}
\begin{table*}[t]
\centering
\begin{tabular}{|c|c|c|c|c|c|c|}
\hline
Normalization scheme & \multicolumn{2}{c|}{Non-adaptive} & \multicolumn{4}{c|}{Adaptive}                                           \\ \cline{1-7}
\textbf{}            Averaging scheme& \multicolumn{2}{c|}{Batch based}  & \multicolumn{2}{c|}{Instance based} & \multicolumn{2}{c|}{Batch based}                                                                                            \\ \cline{1-7}
Statistics           & $\mu,\sigma$      & $\nu^2$     & $\mu,\sigma$       & $\nu^2$      & $\mu,\sigma$     & $\nu^2$                                                                                            \\ \hline
Layer - 24             & 49.84               & 54.93      & 33.24       & 49.61        &  47.65             & 54.20     \\ 
Layer - 36            & 32.41               & 36.60       & 27.25      & 33.19        & 47.73    & 46.48        \\ 
Layer - 44 (Penultimate Layer)          & 24.60      & 31.34       & 26.0                 & 28.26         & 48.53              & 50.44      \\ \hline
\end{tabular}

\caption{Accuracy of decoding the identity of each patient for the \textbf{StrokeRehab} dataset from the features at different layers of convolutional networks trained using different normalization schemes. Networks trained with instance-based adaptive normalization have lower decoding accuracy than the other normalization schemes, which suggests higher invariance to the extraneous variable. Additionally, features from deeper layers show more invariance to the extraneous variable. For comparison, the accuracy of randomly guessing the value of the extraneous variable equals 3.125\% for this dataset.}
\label{tab:strokerehab_decoding_acc_layers}
\end{table*}
\endgroup

\begingroup
\renewcommand*{\arraystretch}{1.5}
\begin{table*}[t]
\centering
\begin{tabular}{|c|c|c|c|c|c|c|}
\hline
Normalization scheme & \multicolumn{2}{c|}{Non-adaptive} & \multicolumn{4}{c|}{Adaptive}                                           \\ \cline{1-7}
\textbf{}            Averaging scheme& \multicolumn{2}{c|}{Batch based}  & \multicolumn{2}{c|}{Instance based} & \multicolumn{2}{c|}{Batch based}                                                                                            \\ \cline{1-7}
Statistics           & $\mu,\sigma$      & $\nu^2$     & $\mu,\sigma$       & $\nu^2$      & $\mu,\sigma$     & $\nu^2$                                                                                            \\ \hline
Layer - 10          &       55.56         &    57.463   &   37.8     &  43.28       &  51.23             &  52.93    \\ 
Layer - 40            & 48.86               &  45.91      & 34.32      & 31.86        & 26.23    &   20.72      \\ 
Layer - 49 (Penultimate Layer)           & 50.57               & 54.97       & 21.92                & 30.28        & 16.93    & 24.31       \\ \hline
\end{tabular}

\caption{Accuracy of decoding the type of corruption for \textbf{CIFAR-10C} dataset from the features of the various layers of convolutional networks trained using different normalization schemes. Networks trained with adaptive normalization schemes tend to have lower decoding accuracy than those trained with non-adaptive normalization schemes, which suggests higher invariance to the extraneous variable. Additionally, features from deeper layers show more invariance to the extraneous variable. For comparison, the accuracy of randomly guessing the value of the extraneous variable equals 5.26\% for this dataset.}
\label{tab:cifar10c_decoding_acc_layers}
\end{table*}
\endgroup

\section{Architectural design and training procedure} \label{app:sec:arch_design_train}
\subsection{StrokeRehab (Section \ref{sec:sensor}) and JIGSAWS (Section \ref{sec:jigsaws})}
\label{sec:sensor_architecture}
\begin{figure*}[t]
\centering
\includegraphics[width=0.9\linewidth]{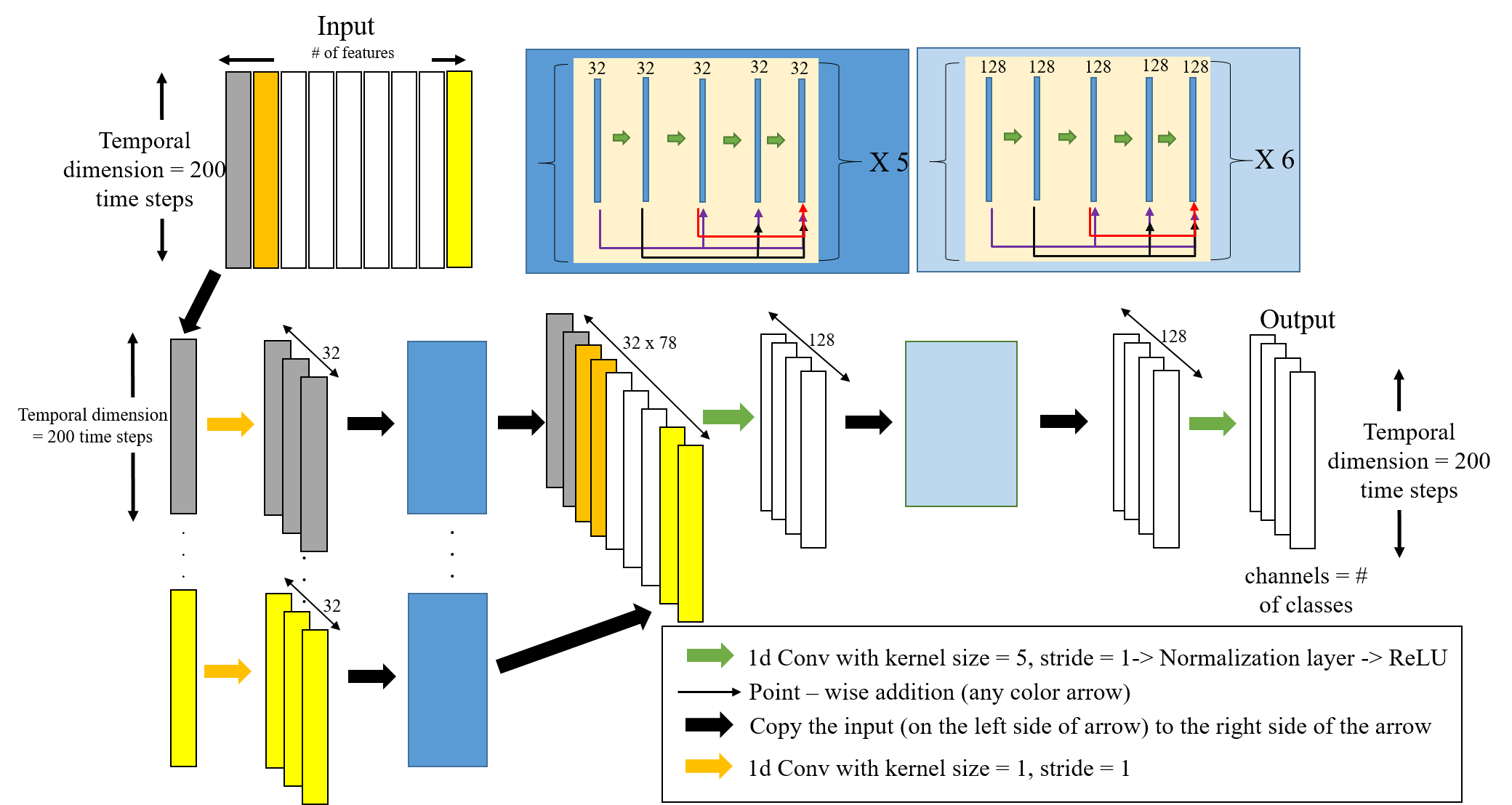}
\caption{DenseNet architecture for JIGSAWS and StrokeRehab dataset as described in Section~\ref{sec:sensor_architecture}. }
\label{fig:dense_net_arch}
\end{figure*}

In this section we describe the modified DenseNet architecture used that we apply on the JIGSAWS and StrokeRehab datasets. The architecture maps sequences of multivariate sensor data to sequences of predictions. The input is a two-dimensional array, where the first dimension corresponds to the different sensor variables ($76$ for JIGSAWS and $78$ for StrokeRehab) and second dimension corresponds to the temporal dimension ($200$ time steps for both datasets). In the first layers of the network, the multi-dimensional time-series is separated into individual time-series and independently passed through $5$ dense blocks that produce separate embeddings for each of them. Then, these embeddings are concatenated and passed through 6 dense blocks. Each dense block consists a set of 4 convolution layers (with 32 or 128 filters), normalization layer and ReLU non linearity. The output of the model has temporal dimension equal to 200 and number of filters equal to the number of classes for classification. A diagram of the network is shown in Figure~\ref{fig:dense_net_arch}.

The procedure applied to train this model was the following:
\begin{enumerate}[topsep=0pt,itemsep=-0.5ex,partopsep=1ex,parsep=2ex]
    \item We used cross-entropy loss as our loss function and the Adam optimizer.
    \item The model is trained for 50 epochs. For the first 35 epochs, the learning rate is set to 5e-4. For the last 15 epochs, the learning rate is set to 5e-5.
    \item Early stopping using validation data is applied to select the model.
\end{enumerate}

\subsection{CIFAR10, CIFAR10-C (Section 5.3) and Fashion-MNIST}

For the image datasets (CIFAR10, CIFAR10-C and Fashion-MNIST) we apply some of the most popular architectures: ResNet50, VGG19 and DenseNet121 on CIFAR10. The architectures of ResNet50 and DenseNet121 contain batch normalization layers. We used a version of VGG19 with batch normalization layers after every convolutional layer \footnote{\url{https://pytorch.org/docs/stable/torchvision/models.html\#torchvision.models.vgg19_bn}}. To create the versions of network with other normalization layers, we just swapped the batch norm layer in the original network with the normalization layer of interest.

\section{Architectural design and training procedure for invariance analysis} \label{app:sec:train_details_invariance_analysis}

To perform the invariance analysis in section \ref{sec:invariance_analysis}, we use a small neural network to decode extraneous variables from features of networks trained with different normalization schemes. The data to train the decoding network was created by pooling together the training, validation, and test sets to obtain the features. Then this set was separated randomly into a train ($60\%), validation (20\%$), and test ($20\%$) set to train and evaluate the decoding network. 

The architecture selected for the decoding network depended on the different datasets. All networks are trained using cross entropy.
\begin{itemize}
    \item \textbf{JIGSAWS} and \textbf{StrokeRehab}: The intermediate layers considered in Tables~\ref{tab:jigsaws_multi_split} and \ref{tab:strokerehab_decoding_acc_layers} have dimensions $128 \times 200$. The network has four convolutional layers with each with $64$ filters, followed by $3$ fully connected layers.
    \item \textbf{CIFAR10-C}: Layers $10$ and $40$,considered in Table~\ref{tab:cifar10c_decoding_acc_layers}, have dimensions $256 \times 32 \times 32$ and $1024 \times 8 \times 8$, respectively. The network has two convolutional layers, followed by three fully connected layers. For the penultimate layer ($2048$ dimensional), we used a fully connected network with $5$ layers.
\end{itemize}

We trained all the networks using a cross entropy loss with the Adam optimizer. The learning rate was decreased if the accuracy on validation set plateaued. We employed early stopping using the accuracy on the validation set to select our final model.

\section{StrokeRehab dataset details}
The StrokeRehab dataset contains data from 32 stroke patients performing nine activities of daily living: feeding, drinking, washing face, brushing teeth, donning glasses, applying deodorant, combing hair, moving objects in radial manner and moving objects on a shelf. Each activity was repeated five times. The approximate duration of activity ranged from 15 secs (for the applying deodorant activity) to 200 secs (for the feeding activity).





\end{document}